\documentclass[letterpaper, 10 pt, conference]{ieeeconf}  

\IEEEoverridecommandlockouts                              

\usepackage{xcolor}
\usepackage{amsmath}
\usepackage{amssymb}
\usepackage{dsfont}
\usepackage{graphicx}
\usepackage{algorithm}
\usepackage{algpseudocode}
\usepackage{romannum}
\usepackage{hyperref}
\usepackage[export]{adjustbox}
\usepackage{subcaption}
\usepackage{pdfpages}

\DeclareFontFamily{OT1}{pzc}{}
\DeclareFontShape{OT1}{pzc}{m}{it}{<-> s * [1.10] pzcmi7t}{}
\DeclareMathAlphabet{\mathpzc}{OT1}{pzc}{m}{it}

\newcommand{\revised}[1]{#1}

\graphicspath{{./figs/}{./figs/sim/}{./figs/real/}}

\title{\LARGE \bf
Attribute-Based Robotic Grasping with One-Grasp Adaptation
}

\author{Yang Yang, Yuanhao Liu, Hengyue Liang, Xibai Lou and Changhyun Choi
\thanks{*This work was supported by \revised{UMII MnDRIVE Ph.D. Graduate Assistantship} and MnDRIVE Initiative on Robotics, Sensors, and Advanced Manufacturing.}
\thanks{$^\dagger$The authors are with University of Minnesota, Minneapolis, USA {\tt\small \{yang5276, liu00800, liang656, lou00015, cchoi\}@umn.edu}}%
}

\begin{document}

\maketitle

\begin{abstract}
Robotic grasping is one of the most fundamental robotic manipulation tasks and has been actively studied. However, how to quickly teach a robot to grasp a novel target object in clutter remains challenging. This paper attempts to tackle the challenge by leveraging object attributes that facilitate recognition, grasping, and quick adaptation. In this work, we introduce an end-to-end learning method of attribute-based robotic grasping with one-grasp adaptation capability. Our approach fuses the embeddings of a workspace image and a query text using a gated-attention mechanism and learns to predict instance grasping affordances. Besides, we utilize object persistence before and after grasping to learn a joint metric space of visual and textual attributes. Our model is self-supervised in a simulation that only uses basic objects of various colors and shapes but generalizes to novel objects and real-world scenes. We further demonstrate that our model is capable of adapting to novel objects with only one grasp data and improving instance grasping performance significantly. Experimental results in both simulation and the real world demonstrate that our approach achieves over 80\% instance grasping success rate on unknown objects, which outperforms several baselines by large margins. Supplementary material is available at \href{https://sites.google.com/umn.edu/attributes-grasping}{https://sites.google.com/umn.edu/attributes-grasping}.
\end{abstract}


\section{INTRODUCTION}
\begin{figure}[t]
  \begin{subfigure}{0.24\textwidth}
    \includegraphics[width=\textwidth]{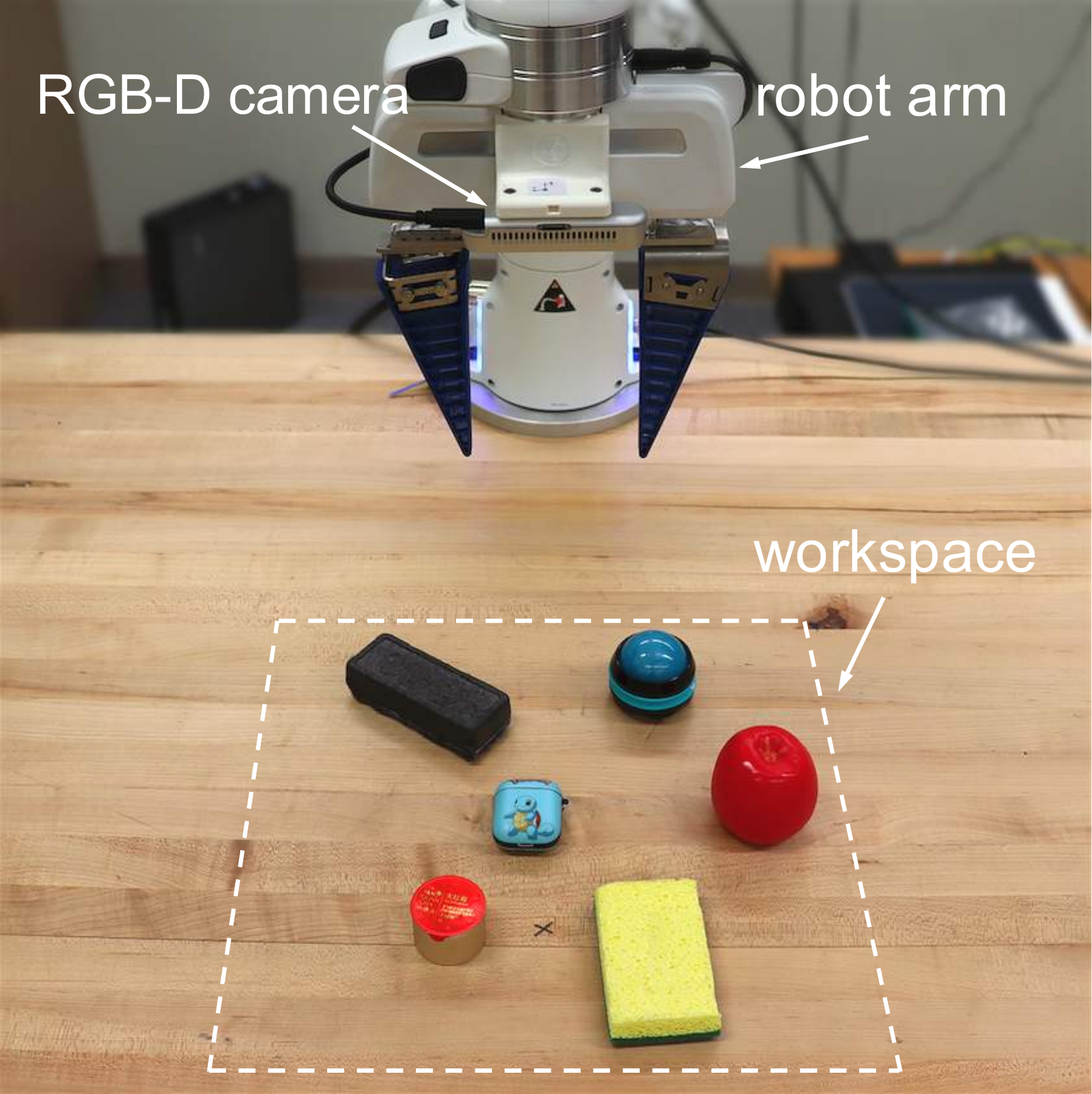}
    \vspace{-10pt}
    \caption{Objects of various attributes}
    \label{fig:intro_a}
  \end{subfigure}
  \hfill
  \begin{subfigure}{0.24\textwidth}
    \includegraphics[width=\textwidth]{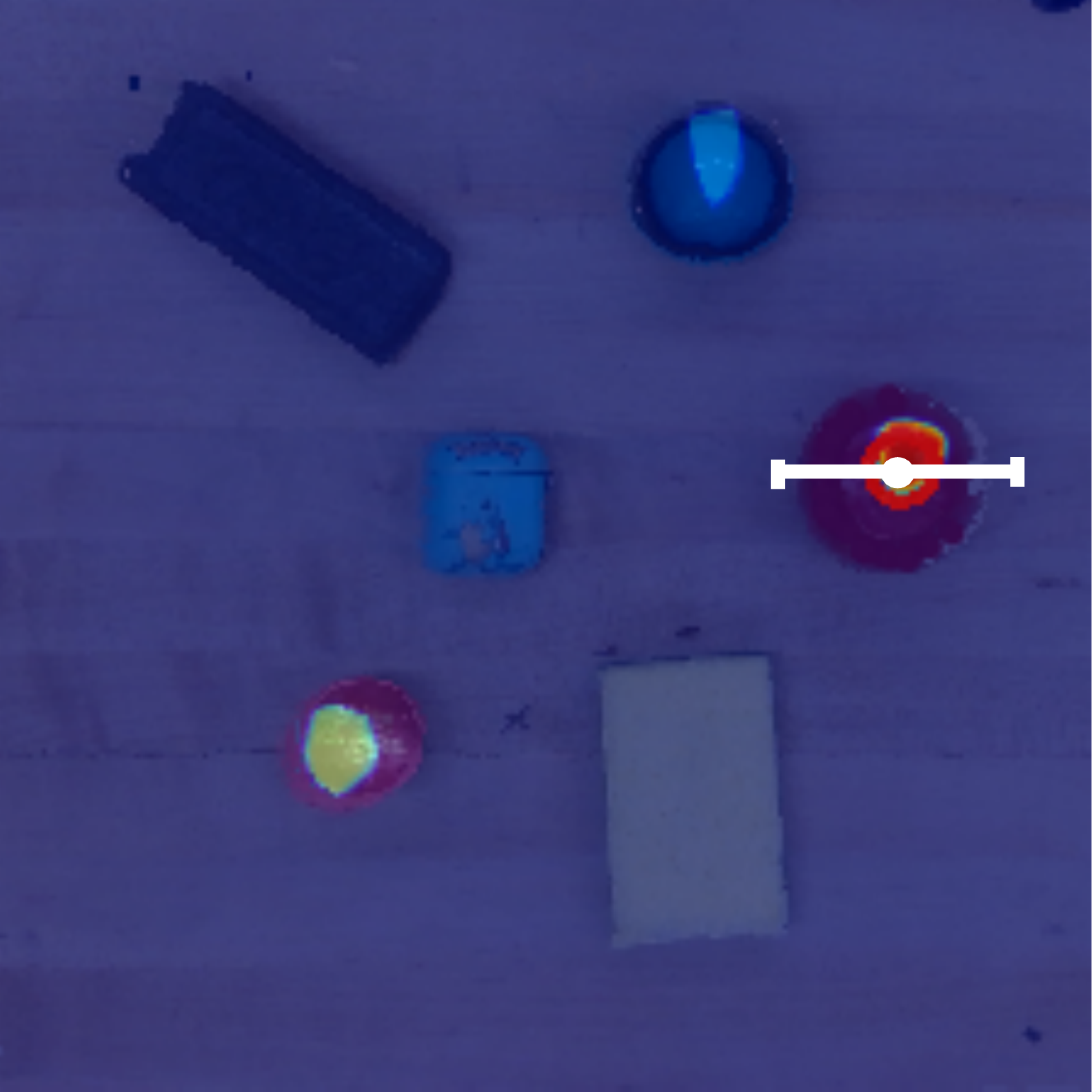}
    \vspace{-10pt}
    \caption{Grasping affordances}
  \end{subfigure}
  \vspace{-3pt}
  \caption{\textbf{Attribute-based instance grasping}. Various objects of generic attributes are placed in the workspace, and we propose to grasp a target object by describing its attributes, e.g., ``apple, red sphere'' for the apple.}
  \label{fig:intro}
  \vspace{-10pt}
\end{figure}

Object attributes are generalizable properties in object manipulation. Imagine how we describe a novel object when asking someone to fetch it, ``give me the apple, a red sphere'', we intuitively describe the target by its appearance attributes (see Fig. \ref{fig:intro}). If an assistive robot can be similarly commanded using such object attributes, it would allow more generalization for novel objects than a discrete set of pre-defined labels. Moreover, humans learn to recognize and grasp an unknown object by quick interactions; it would be beneficial if a grasping pipeline is capable of adapting with minimal adaptation data. These motivate attribute-based robotic grasping with data-efficient adaptation capability.

Recognizing and grasping a target object in clutter is crucial for an autonomous robot to perform everyday-life tasks. Over the last years, the robotics community has made substantial progress in target-driven robotic grasping by combining off-the-shelf object recognition modules with data-driven grasping models \cite{fang2018multi}, \cite{yang2020deep}. However, these recognition-based approaches assume a unique ID for each category and generalize poorly to novel objects. In contrast to applying object recognition modules for robotic grasping, we investigate attribute-based robotic grasping. The intuition of using attributes for grasping is that the grounded attributes can help transfer object recognition and grasping capabilities.

Compared to recognition-based robotic grasping, the challenges of attribute-based grasping are 1) mapping from workspace images and query text describing the target object to robot motions, 2) associating abstract attributes and raw pixels, 3) data labeling in target-driven grasping, and 4) data-efficient adaptation for unknown objects and new scenes.
In this paper, we design an architecture that comprises a multimodal encoder (i.e., encoding visual and textual data) and an affordances decoder (i.e., predicting instance grasping affordances \cite{zeng2019learning}). The key aspects of our system are
\begin{itemize}
    \item The deep neural networks encode and fuse visual-textual representations, rotate the fused representation, and predict pixel-wise instance grasping affordances.
    \item We learn a multimodal metric space by utilizing the equation of object persistence before and after grasping: the visual embedding of a grasped object should be equal to its textual embedding.
    \item Our model learns object attributes that generalize to unknown objects and scenes by only using basic objects (of various colors and shapes) in simulation.
    \item Our approach is capable of further adapting to novel objects and real-world scenes with only one grasp.
\end{itemize}
Our approach is fully self-supervised through robot-object interactions. Fig. \ref{fig:intro} presents an example of attribute-based robotic grasping, wherein our approach successfully disentangles abstract attributes and accurately predicts grasping affordances for the target object apple.

\textbf{Contributions}: This paper presents three core technical contributions: 1) an end-to-end architecture for learning text-commanded robotic manipulation; 2) a method of supervising multimodal embeddings through object grasping, which enables metric learning of object attributes; 3) a one-grasp adaptation scheme \revised{that} only requires one grasp data of a solely placed target object. Our system can perform attributes-specified instance grasping with observations from an RGB-D camera.

\section{RELATED WORK}
\textbf{Instance Grasping}: Though there are different taxonomies, the existing works of robotic grasping can be roughly divided according to approaches and tasks: 1) model-driven \cite{sahbani2012overview} and data-driven \cite{bohg2013data} approaches; 2) indiscriminate \cite{lou2020learning} \cite{zeng2018learning} and instance grasping \cite{fang2018multi} tasks. Our approach is data-driven and focuses on instance grasping. Typical instance grasping pipelines assume a pre-trained object recognition module (e.g., detection \cite{fang2018multi}, segmentation \cite{yang2020deep}, template matching \cite{danielczuk2019mechanical} and object representation \cite{jang2018grasp2vec}, etc.), limiting the generalization for unknown objects and the scalability of grasping pipelines. \revised{Our model is end-to-end and exploits object attributes for generalization. Some recent works also propose to learn instance grasping end-to-end. \cite{jang2017end} learns to predict the grasp configuration for an object class with a learning framework of object detection, classification, and grasp planning.}
In \cite{cai2020ccan}, CCAN, an attention-based network, learns to locate the target object and predict corresponding grasp affordance given a query image. \revised{Compared to these methods, the main features of our work are two-fold. First, we collect a much smaller dataset of synthetic basic objects to learn generic attribute-based grasping. Moreover, our generic grasping model is capable of further adapting to unknown objects and domains. Second, our approach takes a text description of target attributes as a query command, which is more flexible when grasping a novel object.}

\textbf{Attribute-Based Methods}: Object attributes are \revised{middle}-level abstractions of object properties and generalizable across object categories \cite{farhadi2009describing}. Learning object attributes has been widely studied in tasks of object recognition \cite{zhong2000object}, \cite{sun2013attribute}, \cite{hermans2011affordance}, \cite{pirk2020online}, while attribute-based robotic grasping has been much less explored, except for \cite{cohen2019grounding}, \cite{ahn2018interactive}. Cohen \emph{et al.} \cite{cohen2019grounding} developed a robotic system to pick up the target object corresponding to a description of attributes. Their approach minimizes the cosine similarity loss between visual and textual embeddings as well as predicts object attributes. However, they only show generalization across viewpoints but not object categories. In \cite{ahn2018interactive}, the proposed Text2Pickup system uses object attributes to specify a target object and remove ambiguities in the user’s command. They use mono-color blocks as training and testing objects but fail to show generalization to novel objects. In contrast, our work learns generic attribute-based robotic grasping (only using synthetic basic objects) and generalizes well to novel objects and real-world scenes.

\textbf{Few-Shot Learning and Adaptation}: Few-shot learning \cite{fei2006one} is the paradigm of learning from a small number of examples at test time. The key of metric-based few-shot learning methods, one of the most popular categories, is to supervise a latent space and learn a versatile similarity function by metric loss \cite{koch2015siamese}, \cite{snell2017prototypical}. The supervised metric space supports fine-tuning and adaptation using a minimal adaptation data (also known as support set), and the similarity function generalizes to unknown test data \cite{motiian2017few}, \cite{dhillon2019baseline}. Motivated by the idea of the few-shot learning methods, our approach first learns a joint metric space that encodes object attributes and then fine-tunes recognition and grasping of our model when testing on novel objects.

\section{PROBLEM FORMULATION}
The attribute-based robotic grasping problem in this paper is formulated as follows:

\vspace{3pt}\noindent\textbf{Definition 1.} \textit{Given a query text for a target object, the goal is to grasp the corresponding object that is placed in the workspace with clutter.}\vspace{3pt}

We consider color, shape, and name attributes in this paper and leave other attributes (e.g., texture, pose, and functionality, etc.) as future work. To make object recognition tractable, we have an assumption about object placement:

\vspace{3pt}\noindent\textbf{Assumption 1.} \textit{The objects are stably placed within the workspace, and there is no stacking between objects.}




While we show robotic grasping as a manipulation example in this paper, the proposed attribute-based learning methods should be, in principle, extendable to other robotic manipulation skills, such as pushing and suction.

\section{METHODS}
\subsection{Learning Grasping Affordances}
\label{subsec:method_grasp}
\begin{figure*}[t]
  \centering
  \includegraphics[width=\textwidth]{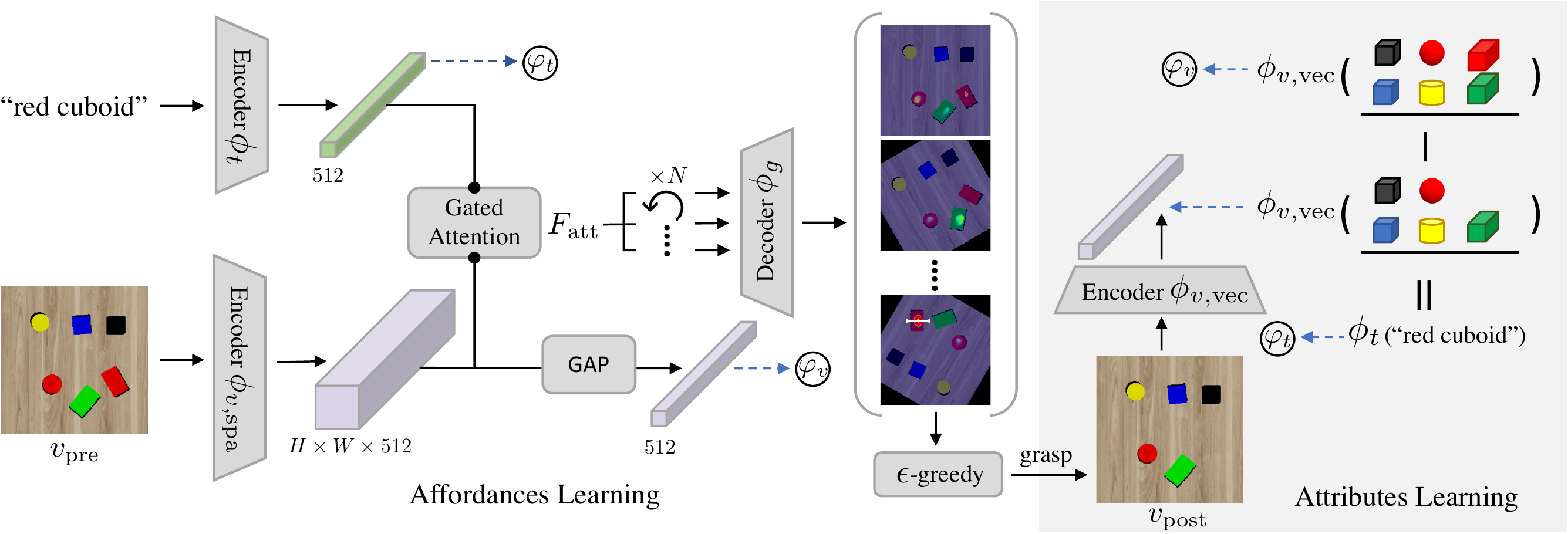}
  \vspace{-16pt}
  \caption{\textbf{Overview of learning.} The workspace image and query text are encoded separately and fused using gated-attention. The fusion matrix $F_{\text{att}}$ is rotated by $N$ orientations for different grasping angles and then fed into the grasping affordances decoder. The decoder learns to predict pixel-wise scores of target grasping success, and we run the $\epsilon$-greedy grasping policy and obtain the image $v_{\text{post}}$ after grasping. By utilizing the equation of object persistence before and after grasping, we learn a metric space where multimodal embedding vectors corresponding to similar attributes are encouraged to be closer. Note we denote the combination of $\phi_{v, \text{spa}}$ and GAP as $\phi_{v, \text{vec}}$.}
  \vspace{-10pt}
  \label{fig:overview}
\end{figure*}

We formulate attribute-based grasping as a mapping from workspace images and query text to target grasping affordances. The proposed visual-text manipulation architecture assumes no prior linguistic or perceptual knowledge. It consists of two modules, a multimodal encoder and an affordances decoder, as illustrated in Fig. \ref{fig:overview}.

\textbf{Multimodal Encoder}: An overhead RGB-D camera captures the predefined workspace. Then the images are orthographically projected in the gravity direction to construct RGB-D heightmap $v_{\text{pre}}$. To specify the target to be grasped, we give a text command $t$ composed of color and/or shape attributes, e.g., ``red cuboid''. The workspace image $v_{\text{pre}}$ and query text $t$ are the input to visual spatial encoder $\phi_{v, \text{spa}}$ and text encoder $\phi_t$ respectively. We use ImageNet-pretrained \cite{deng2009imagenet} ResNet-18 \cite{he2016deep} as the encoder $\phi_{v, \text{spa}}$ to produce 3D visual matrix $\varphi_{v, \text{spa}} \in \mathbb{R}^{H \times W\times 512}$. The text encoder $\phi_t$ is a \revised{deep averaging network \cite{iyyer2015deep}} that takes as input the mean token embeddings of all the tokens in a sentence and produces a text vector $\varphi_{t} \in \mathbb{R}^{512}$. The visual matrix $\varphi_{v, \text{spa}}$ and the text vector $\varphi_{t}$ are then fused by the gated-attention mechanism \cite{chaplot2018gated}: each element of $\varphi_{t}$ is repeated and expanded to an $H \times W$ matrix to match the dimension of $\varphi_{v, \text{spa}}$. The expanded matrix is multiplied element-wise with $\varphi_{v, \text{spa}}$ to produce a fusion matrix $F_\text{att}$, which contains selected visual features by the query text.

\textbf{Affordances Decoder}: Grasping affordances decoder $\phi_g$ is a fully-convolutional residual network \cite{he2016deep}, \cite{long2015fully} interleaved with spatial bilinear upsampling and ended with the sigmoid function. The decoder takes as input the fusion matrix $F_\text{att}$ and outputs a unit-ranged map $Q_g$ with the same size as the input image $v_{\text{pre}}$. Each value of a pixel $q_i \in Q_g$ represents the predicted score of target grasping success when executing a top-down grasp at the corresponding 3D location with a parallel-jaw gripper oriented horizontally concerning the map $Q_g$. Different grasping angles are examined by rotating $F_\text{att}$ by $N = 6$ (multiples of $30^{\circ}$) orientations before feeding into the decoder. The pixel with the highest score among all the $N$ maps determines the parameters for the grasping primitive to be executed. As in Fig. \ref{fig:overview}, our model predicts accurate target grasping location and valid (e.g., the selected angles for the red cuboid) target grasping angle.

The motion loss $\mathcal{L}_m$, which supervises the entire networks, is the error from predictions of grasping affordances:
\begin{align}
    \label{eq:motion}
    \mathcal{L}_m = (q_e-\bar{q}_e)^2 + \frac{\lambda_M}{|M|} \sum_{i \in M} q_i^2
\end{align}
where $q_e$ is the grasping score in $Q_g$ at the executed location, and $\bar{q}_e$ is the ground-truth label (see Sec. \ref{subsec:method_data}). The second term ensures a zero grasping score for the pixels in background mask $M$ (from the depth image) with weight $\lambda_M$ \cite{liang2019knowledge}, and $q_i$ is the grasping score of a background pixel.
\subsection{Learning Multimodal Attributes} \label{subsec:method_metric}
To learn generic object attributes, we perform multimodal attributes learning, where visual or textual embedding vectors corresponding to similar attributes are encouraged to be closer in the latent space. Inspired by \cite{jang2018grasp2vec}, we take advantage of the interactions between the robot and objects: the embedding difference of the scene before and after grasping is enforced closer to the representation of the grasped object. During data collection, we record visual-text triplets $(v_\text{pre}, v_\text{post}, t)$, where $v_\text{pre}$ and $v_\text{post}$ are the workspace image before and after grasping respectively, and $t$ is the query text that describes attributes of the grasped object.

We add one layer of global average pooling (GAP) \cite{lin2013network}, \cite{zhou2016learning} at the end of the encoder $\phi_{v, \text{spa}}$ and denote the network as visual vector encoder $\phi_{v, \text{vec}}$. The output from $\phi_{v, \text{vec}}$ is a visual embedding vector that represents the average of scene features and has the same dimension of $\phi_t(t)$. We express the logic of object persistence (in the previous paragraph) as an arithmetic constraint on visual and textual vectors such that $(\phi_{v, \text{vec}}(v_\text{pre}) - \phi_{v, \text{vec}}(v_\text{post}))$ is close to $\phi_t(t)$. We use the triplet loss \cite{schroff2015facenet} to approximate the constraint, and the set of triplets $\mathcal{T}$ is defined as
\begin{align}
    \mathcal{T} = \left\{(f_i, f_i^+, f_i^-) \mid s(\mathpzc{a}_{f_i}, \mathpzc{a}_{f_i^+}) > s(\mathpzc{a}_{f_i}, \mathpzc{a}_{f_i^-})\right\}
\end{align}
where $f_i$, $f_i^+$ and $f_i^-$ are \revised{random} samples from the pool of vectors $(\phi_{v, \text{vec}}(v_\text{pre}) - \phi_{v, \text{vec}}(v_\text{post}))$ and $\phi_t(t)$, and $\mathpzc{a}_{f}$ is an $n$-dimensional attribute label vector corresponding to vector $f$. Function $s(\cdot, \cdot)$ is an attribute similarity function that evaluates the similarity between two attribute label vectors:
\begin{align}
    \label{eq:similarity}
    s(\mathpzc{a}_1, \mathpzc{a}_2) &= \frac{1}{n}\sum_{i=1}^n \mathds{1}(\mathpzc{a}_1^i, \mathpzc{a}_2^i)\\
    \mathds{1}(\mathpzc{a}_1^i, \mathpzc{a}_2^i) &= \begin{cases} 
      1 & \text{if}~\mathpzc{a}_1^i = \mathpzc{a}_2^i \neq 0 \\
      0 & \text{otherwise}
    \end{cases}
\end{align}
where $\mathpzc{a}^i$ denotes the $i$-th element of the label vector $\mathpzc{a}$, and the indicator function $\mathds{1}(\cdot, \cdot)$ evaluates the element-wise similarity. Note that $0$ indicates null attribute meaning no attribute is specified in the label (e.g., $s((1, 3)^\top, (2, 3)^\top) = 0.5$ and $s((1, 0)^\top, (2, 0)^\top) = 0$). With the triplets of embedding vectors, multimodal metric loss $\mathcal{L}_r$ is defined as
\begin{align}
    \label{eq:metric}
    \mathcal{L}_r(\mathcal{T})= \sum_{i=1}^{|\mathcal{T}|} \max \left(\|f_i - f_i^+\|^2-\|f_i- f_i^-\|^2 + \alpha, 0\right)
\end{align}
where $\alpha$ is a hyperparameter that controls the margin between positive and negative pairs. By encoding workspace images and query text into a joint metric space and supervising the embeddings through the equation of object persistence, we learn generic attributes that are consistent across object categories, as discussed in Sec. \ref{subsec:exp_metric}. 
%
\begin{algorithm}[b]
\caption{Online Data Collection}
\textbf{Initialize} bounded buffer $B$\\
\textbf{Notations}: $\epsilon$-greedy policy $\pi_\epsilon$, our model $\phi$, image $v$, text $t$, mask $M$, action $a$, and label $\bar{q}_e$
\begin{algorithmic}[1]
\While{collecting data}
\State reset the simulation and randomly drop basic objects
\State get image $v_{\text{pre}}$ and randomly choose a command $t$
\State execute action $a \gets \pi_{\epsilon}(\phi; v_{\text{pre}}, t, \epsilon)$
\State label $\bar{q}_e$ according to grasping results
\State save $v_{\text{pre}}$, $t$, $M$, $a$ and $\bar{q}_e$ into $B$
\If{successful grasp}
\State save $v_{\text{post}}$ into $B$ with HER
\EndIf
\State sample a batch from $B$ to train the model
\EndWhile
\end{algorithmic}
\label{alg:training}
\end{algorithm}
\begin{figure}[t]
  \begin{subfigure}{0.23\textwidth}
    \includegraphics[width=\textwidth]{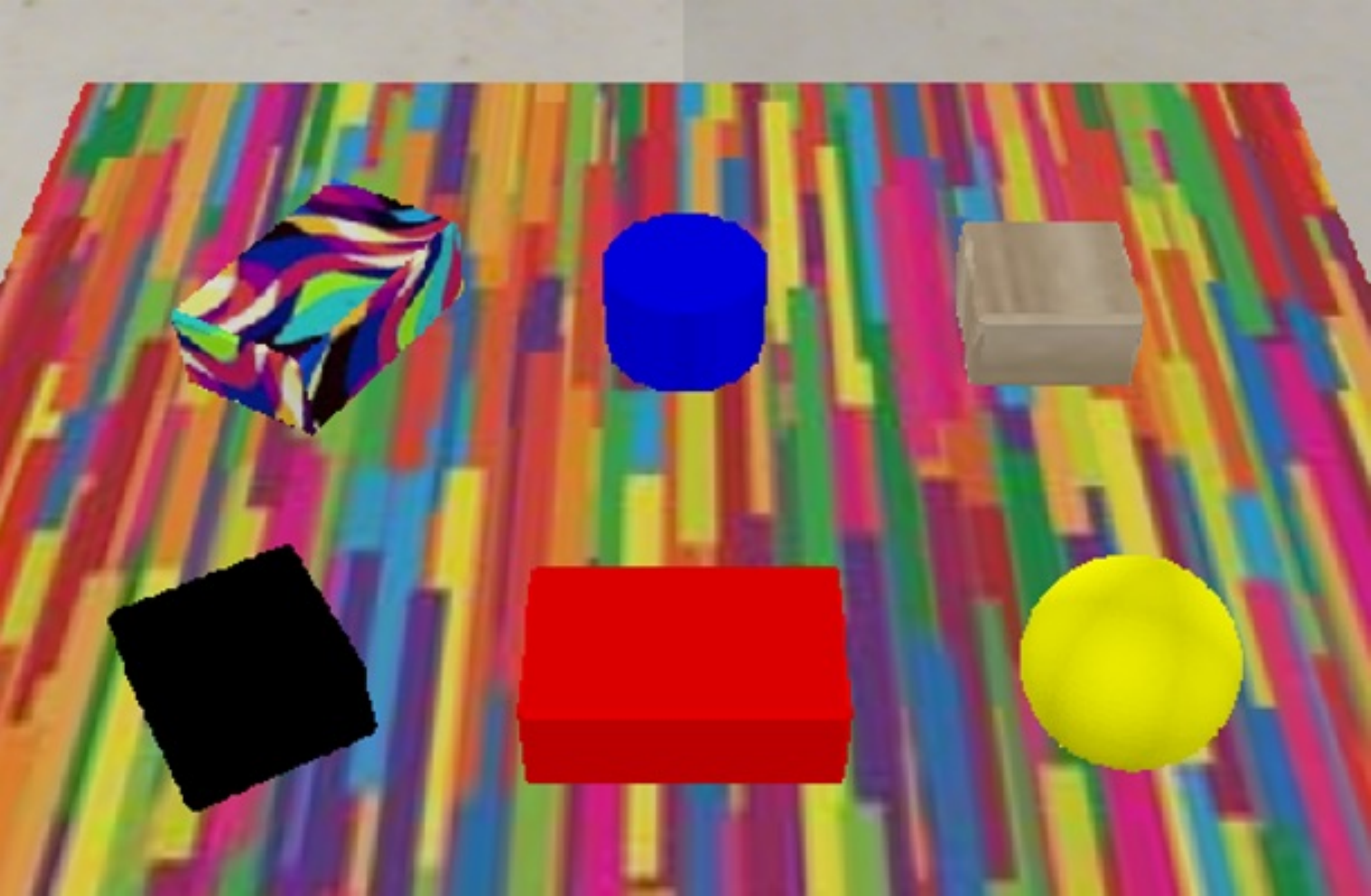}
  \end{subfigure}
  \hfill
  \begin{subfigure}{0.23\textwidth}
    \includegraphics[width=\textwidth]{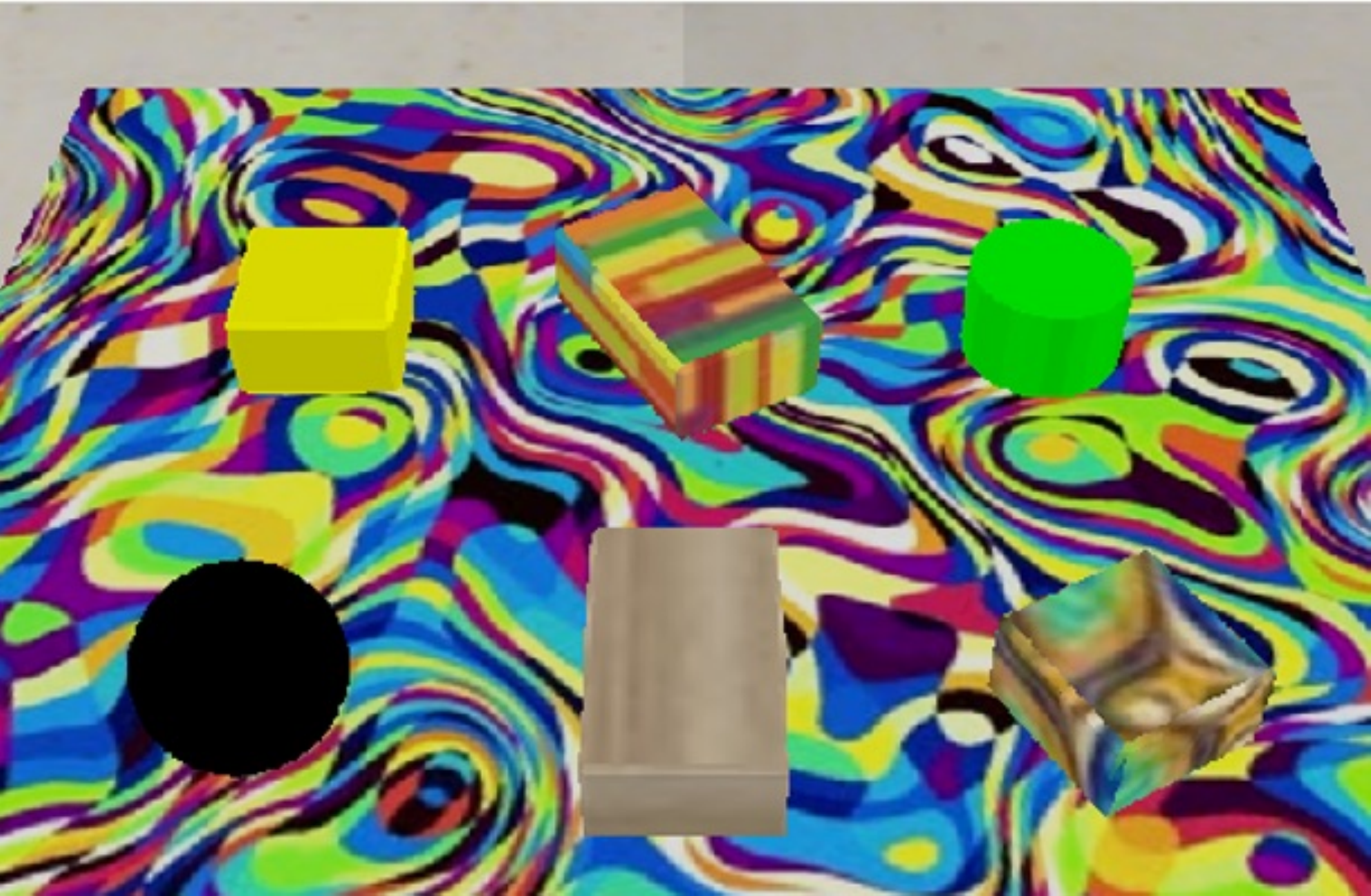}
  \end{subfigure}
  \caption{\textbf{Examples of basic objects}. Synthetic objects of various colors and shapes are used for learning object attributes and grasping affordances. To ensure shape attribute learning, we include objects having random textures.}
  \vspace{-10pt}
  \label{fig:basic}
\end{figure}
\begin{figure}[t]
  \centering
  \begin{subfigure}{0.27\textwidth}
    \includegraphics[width=\textwidth]{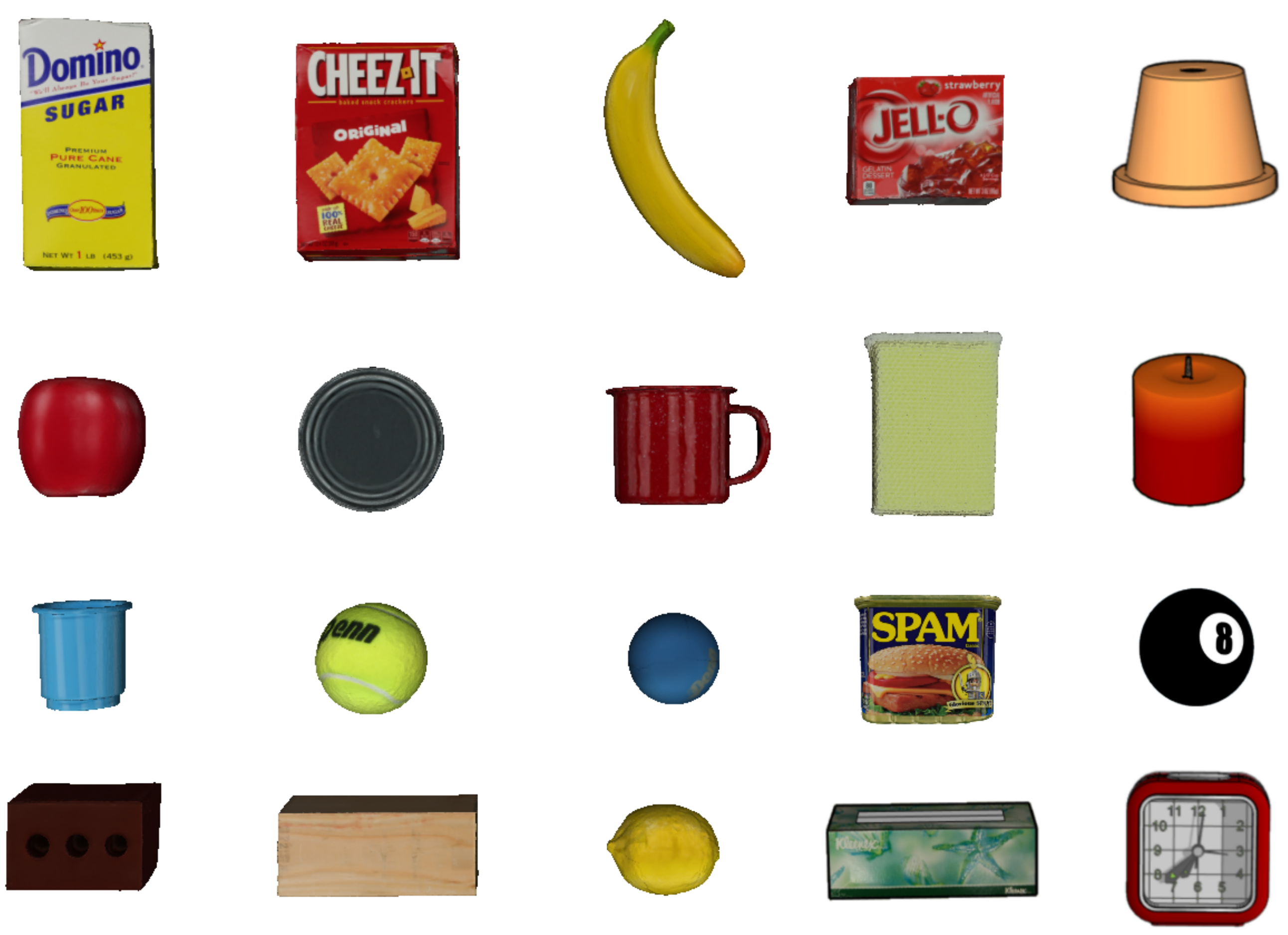}
    \caption{Simulated novel objects}
    \label{fig:testing_sim}
  \end{subfigure}
  \hfill
  \begin{subfigure}{0.2\textwidth}
    \includegraphics[width=\textwidth]{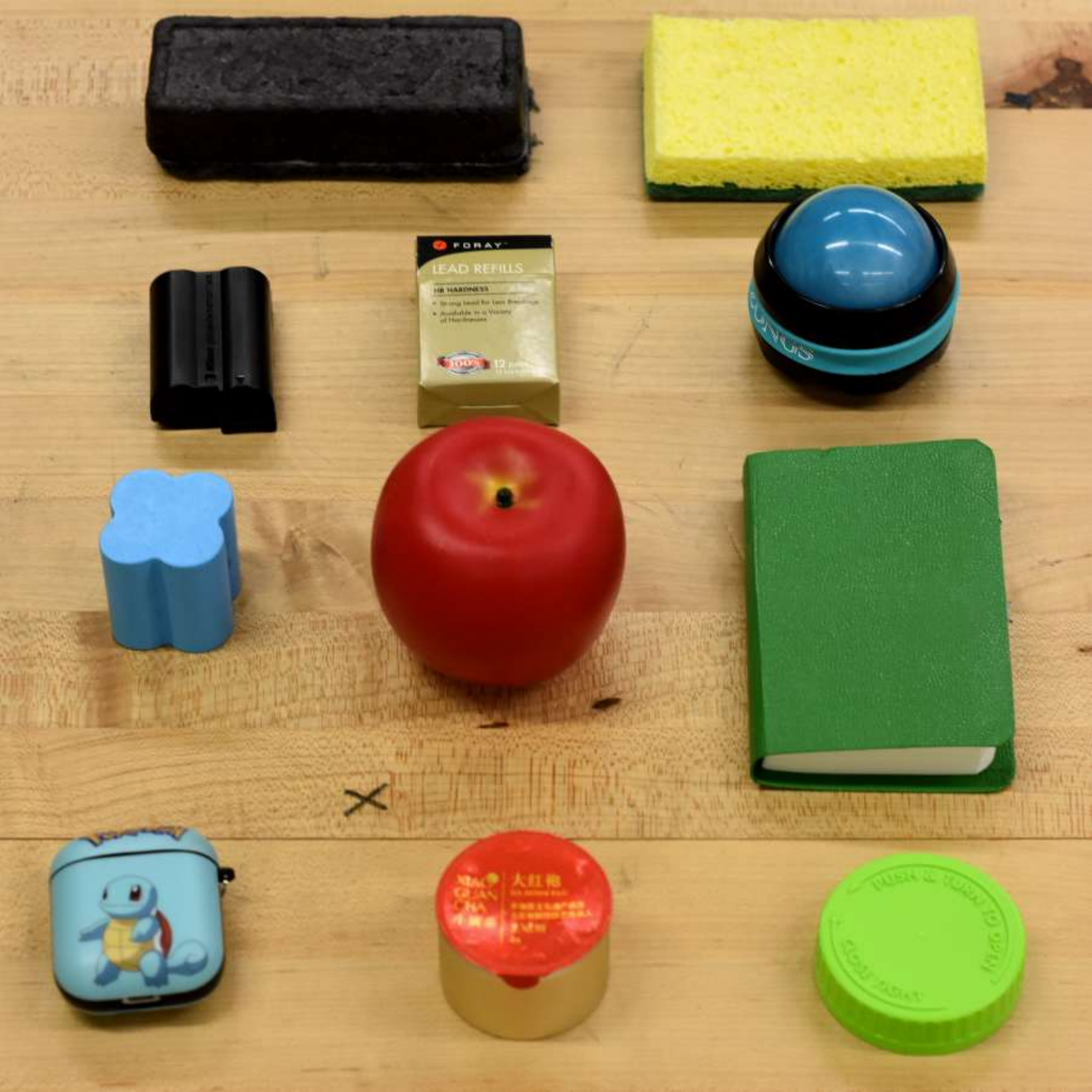}
    \caption{Real-world objects}
    \label{fig:testing_real}
  \end{subfigure}
  \vspace{-3pt}
  \caption{\textbf{Testing objects.} We use the testing objects that share similar attributes with the training objects.}
  \vspace{-10pt}
  \label{fig:testing}
\end{figure}

\subsection{Data Collection and Training} \label{subsec:method_data}
To achieve self-supervision, we collect training data in simulation with the following procedure, as summarized in Algorithm \ref{alg:training}. Several objects are randomly dropped into the workspace in front of the robot. Given a workspace image and a query text, the robot learns to grasp a target under $\epsilon$-greedy exploration \cite{sutton2018reinforcement} ($\epsilon\!=\!0$ during testing, i.e., an argmax policy). We save the workspace images, query text, background masks, executed actions, and results into a bounded buffer. The ground-truth labels are automatically generated for learning grasping affordances. The label $\bar{q}_e$ in (\ref{eq:motion}) is assigned as the attribute similarity in (\ref{eq:similarity}) between the query text and the grasped object (0 if no object grasped). We also save workspace image after a successful grasping for learning object attributes. To deal with sparse rewards in target-driven grasping, the hindsight experience replay (HER) technique \cite{andrychowicz2017hindsight} is applied. Fig. \ref{fig:basic} shows the basic objects of various colors (red, green, blue, yellow, and black) and shapes (cube, cuboid, cylinder, and sphere) used in our simulation. We choose these colors and shapes because they are foundational for common objects in daily life. To enrich the distribution of training data, we perturb color RGB values, randomize sizes and heights of the objects, and randomize textures of the workspace for domain randomization. At every iteration, we sample a batch from the buffer and run one off-policy training. The training loss is defined as
\begin{align}
    \mathcal{L} = \mathcal{L}_m + \lambda_r \mathcal{L}_r
\end{align}
by combining both motion loss $\mathcal{L}_m$ (\ref{eq:motion}) and metric loss $\mathcal{L}_r$ (\ref{eq:metric}). After collecting a dataset of 5k iterations, we replay the entire data for several iterations.
\begin{figure}[b]
  \begin{subfigure}[t]{0.1152\textwidth}
    \includegraphics[width=\textwidth]{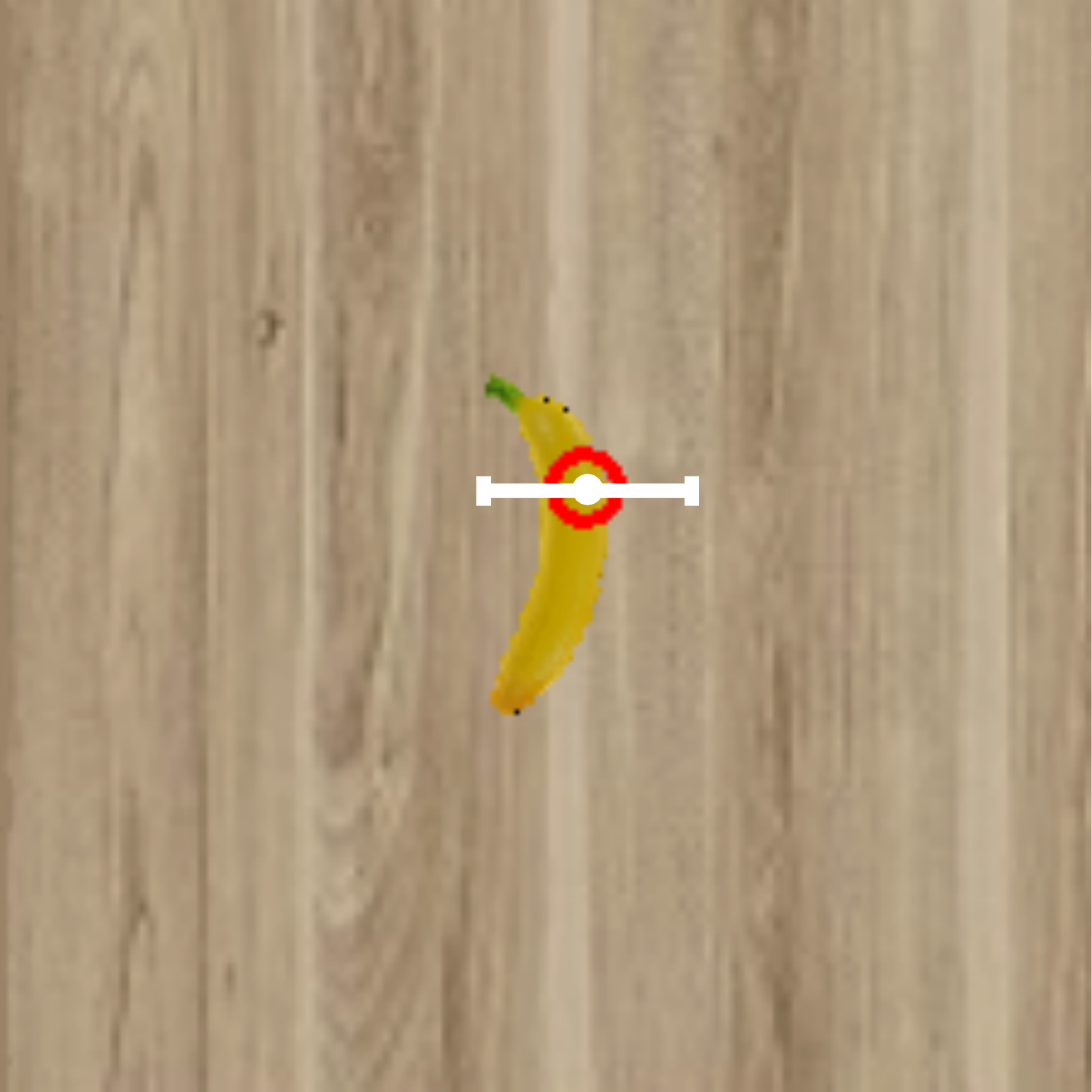}
    \caption*{collected data}
  \end{subfigure}
  \begin{subfigure}[t]{0.1152\textwidth}
    \includegraphics[width=\textwidth]{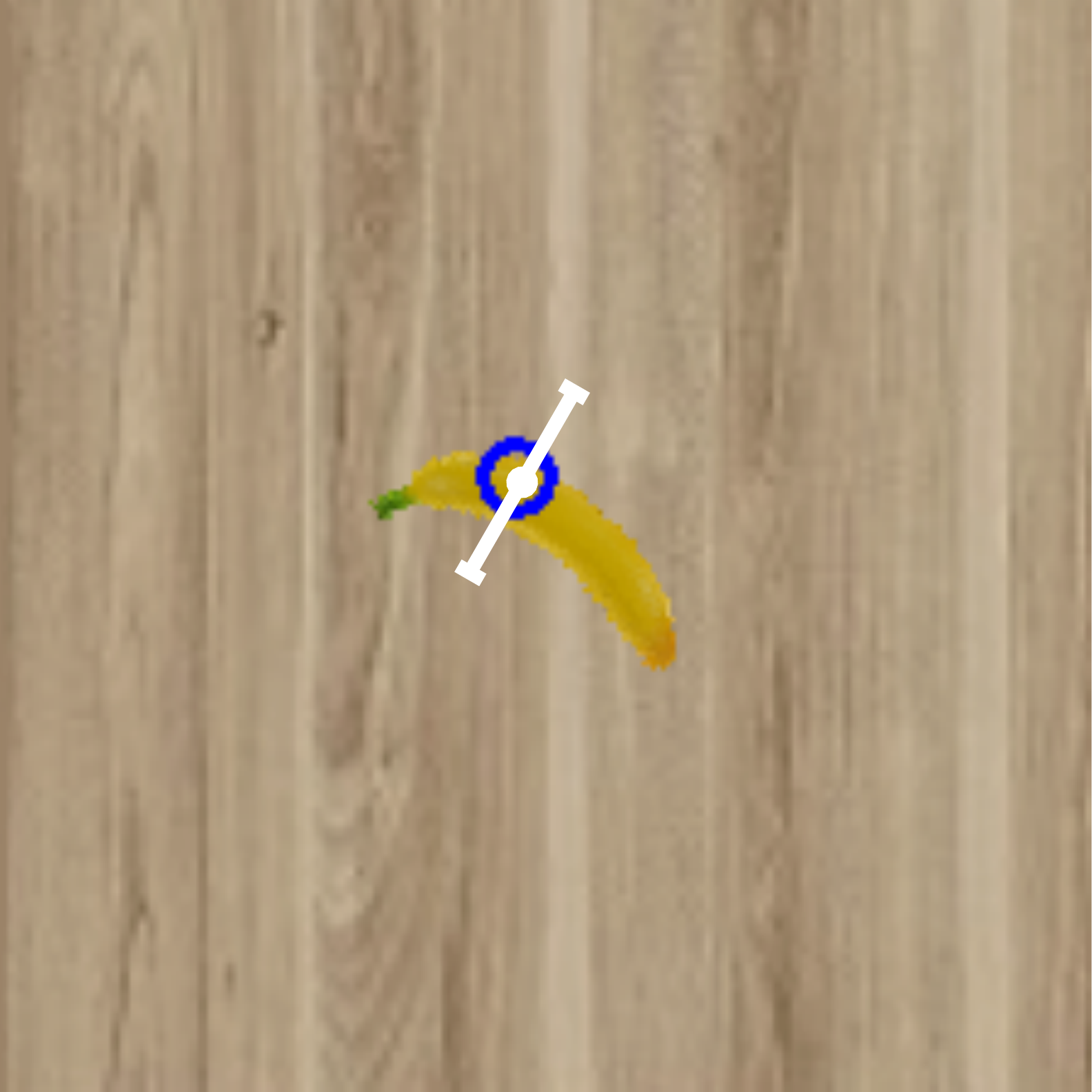}
    \caption*{augmented data}
  \end{subfigure}
  \begin{subfigure}[t]{0.1152\textwidth}
    \includegraphics[width=\textwidth]{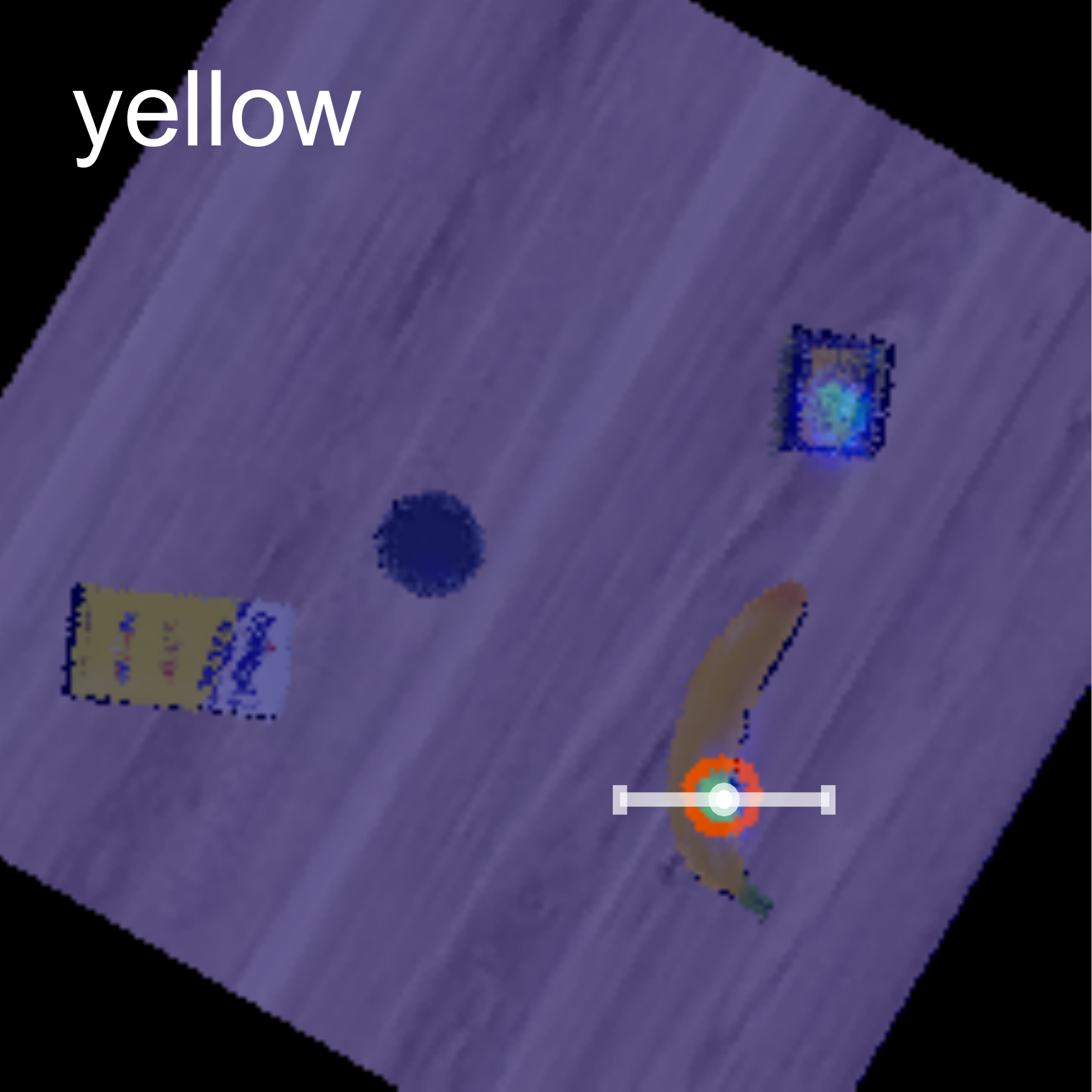}
    \caption*{generic model}
  \end{subfigure}
  \begin{subfigure}[t]{0.1152\textwidth}
    \includegraphics[width=\textwidth]{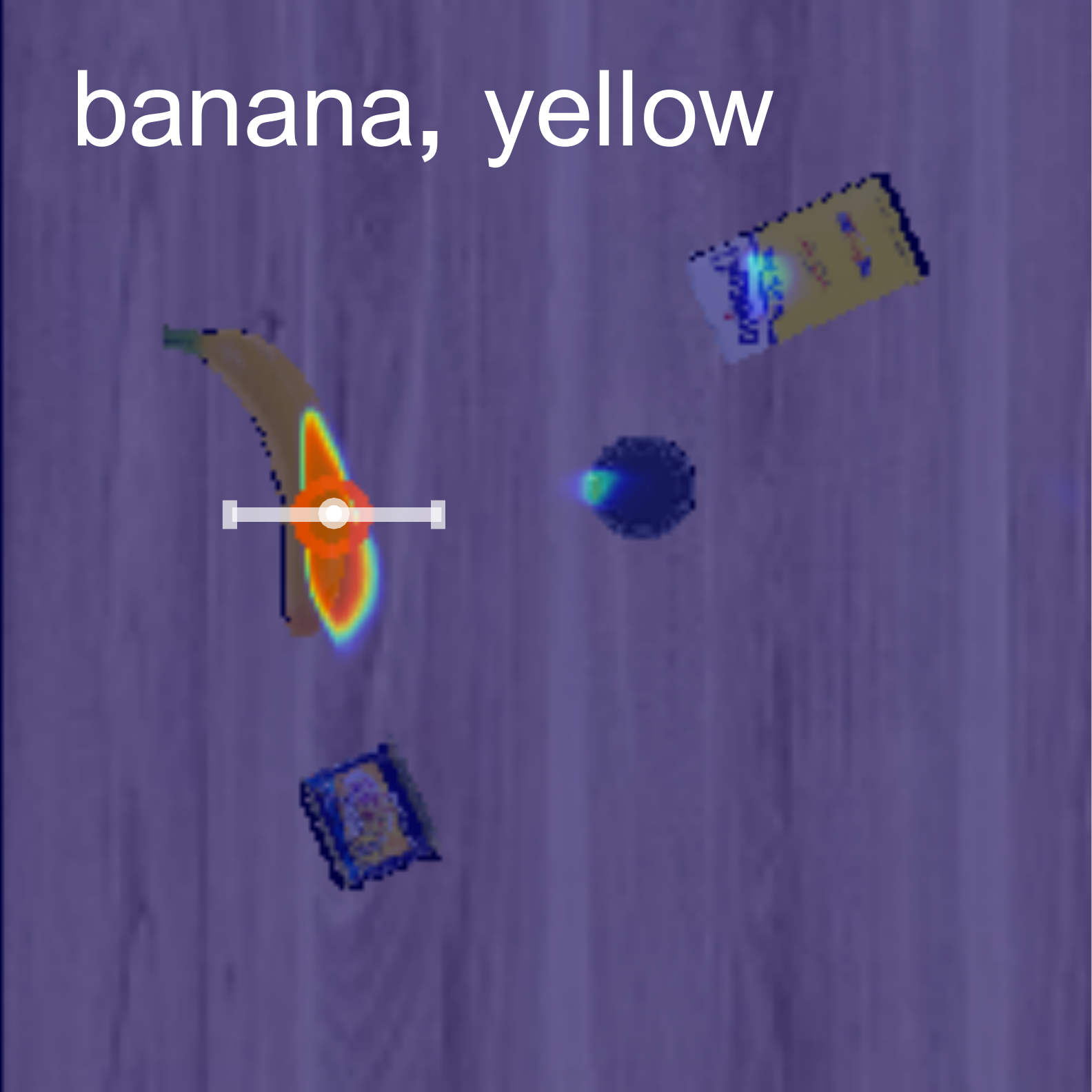}
    \caption*{adapted model}
  \end{subfigure}
  \vspace{-3pt}
  \caption{\textbf{Visualization of the adaptation scheme}. We collect one successful grasp and rotate it by various angles for data augmentation. After being fine-tuned on the adaptation data, the adapted model increases its affordances for the novel target object banana.}
  \label{fig:adapt}
\end{figure}
\subsection{One-Grasp Adaptation}
\label{subsec:method_adapt}
Though our generic model (trained using the simulated basic objects) shows good generalization (see Sec. \ref{subsec:exp_grasp}), the gaps in testing objects and domains are inevitable. To further adapt to novel objects and new scenes, we propose a one-grasp adaptation scheme that only requires one successful grasp of a novel object. The adaptation data is collected with the following procedure. We place the object solely in the workspace and run the generic model to collect one successful grasp. The setting of a sole object facilitates grasping and avoids combinatorial object arrangements. Because convolutional neural networks are not rotation-invariant by design, we also augment the grasp data by rotating with various orientations to achieve rotation-invariance \cite{jaderberg2015spatial}, \cite{quiroga2018revisiting}, i.e., the ability to recognize and grasp an object regardless of its orientation. As shown in Fig. \ref{fig:adapt}, we rotate the collected image and action execution to have rotated versions of the collected data.

In the adaptation stage, we add the name of the object as an additional token to the query text, e.g., ``apple, red sphere'' for the testing object apple. The token embedding of the object name is initialized properly to keep the embedding vector of the query text unchanged. The addition of the object name allows for a more specific grasping instruction and distinguishing from similar objects via adaptation. By optimizing over motion loss $\mathcal{L}_m$ (\ref{eq:motion}), we jointly fine-tune recognition and grasping of our model for unknown objects and scenes. As shown in Fig. \ref{fig:adapt}, the adapted model outputs higher affordances on the YCB object banana that is not seen and grasped before. More details are delineated in Sec. \ref{subsec:exp_adapt}.

\section{EXPERIMENTS}
We first train our approach with simulated basic objects to have a generic model and then adapt it to novel objects and real-world scenes. We analyze the structured metric space of our generic model and evaluate the instance grasping performance of our approach before and after one-grasp adaptation. The goals of the experiments are 1) to show the effectiveness of multimodal attribute learning for instance grasping, 2) to evaluate our attribute-based grasping system in both simulated and real-world settings, and 3) to show the one-grasp adaptation capability of our approach. Our simulation setup uses a UR5 robot arm with an RG2 gripper in CoppeliaSim \cite{rohmer2013v}. Fig. \ref{fig:intro_a} shows our real-world configuration that consists of a Franka Emika Panda robot arm with a FESTO DHAS soft gripper\footnote{We use the soft fingers because they are more suitable for grasping the objects in our experiments and are similarly compliant to the RG2 fingers.} and a hand-mounted Intel RealSense D415 camera overlooking a tabletop scene.

\subsection{Multimodal Attention Analysis}\label{subsec:exp_metric}
By embedding workspace images and query text into a joint metric space, the multimodal encoder ($\phi_v$ and $\phi_t$), supervised by metric loss $\mathcal{L}_r$ and motion loss $\mathcal{L}_m$, learns attending to text-correlated visual features. We visualize what our model ``sees'' by computing the dot product of text vector $\varphi_t$ with each pixel of visual matrix $\varphi_{v, spa}$. This computation obtains an attention heatmap over the image, which refers to the similarity between the query text and each pixel’s receptive field (see Fig. \ref{fig:metric}). We  quantitate the attention of our model and report its attention localization performance in Table \ref{tab:acc} (see \textbf{\textit{Ours-Attention}}). An attention localization is correct if the maximum point in the attention heatmap lies on the target object. Our generic model performs localization at 79.4\% accuracy on simulated novel objects and 75.7\% accuracy on real-world objects, without any localization supervision provided. In summary, our multimodal embeddings demonstrate a consistent pattern across object categories and scenes. Though the localization results are not directly used for grasping, the consistent embeddings facilitate learning, generalization, and adaptation of our grasping model, as shown in Fig. \ref{fig:sim} and discussed in the following subsections.

\subsection{Generic Instance Grasping}\label{subsec:exp_grasp}
We compare the instance grasping performance of our generic model with the following baselines: 1) \textbf{\textit{Indiscriminate}} is an indiscriminate grasping version of our approach and composed of a visual spatial encoder $\phi_{v, \text{spa}}$ and a decoder $\phi_g$. We collect a dataset of binary indiscriminate grasping labels and train \textit{Indiscriminate} using $\mathcal{L}_m$ (\ref{eq:motion}). 2) \textbf{\textit{ClassIndis}} extends \textit{Indiscriminate} with an attributes classifier that is trained to predict color and shape attributes on cropped object images. We filter the grasping maps from \textit{Indiscriminate} using the mask of a recognized target by the classifier. 3) \textbf{\textit{EncoderIndis}} is similar to \cite{cohen2019grounding} and is another extension of \textit{Indiscriminate}, which leverages a multimodal encoder ($\phi_{v, \text{vec}}$ and $\phi_t$ in Sec. \ref{subsec:method_grasp}) for text template matching. The encoder is trained using $\mathcal{L}_r$ (\ref{eq:metric}) to evaluate the similarity between each cropped object image and query text. We also include attributes classification as an axillary task. 4) \textbf{\textit{NoMetric}} is for an ablation study of multimodal metric loss. We simply remove the metric loss on the basis of our approach during its training. These methods have different target recognition schemes: \textit{ClassIndis} and \textit{EncoderIndis} recognize a target by classification and text template matching respectively; \textit{NoMetric} and \textit{Ours} are end-to-end and we report their grasping localization performance (in addition to instance grasping performance). A grasping localization is correct if the predicted grasping location lies on the target object. 
\begin{figure}[b]
    \centering
    \begin{subfigure}{0.48\textwidth}
        {\includegraphics[width=0.24\textwidth]{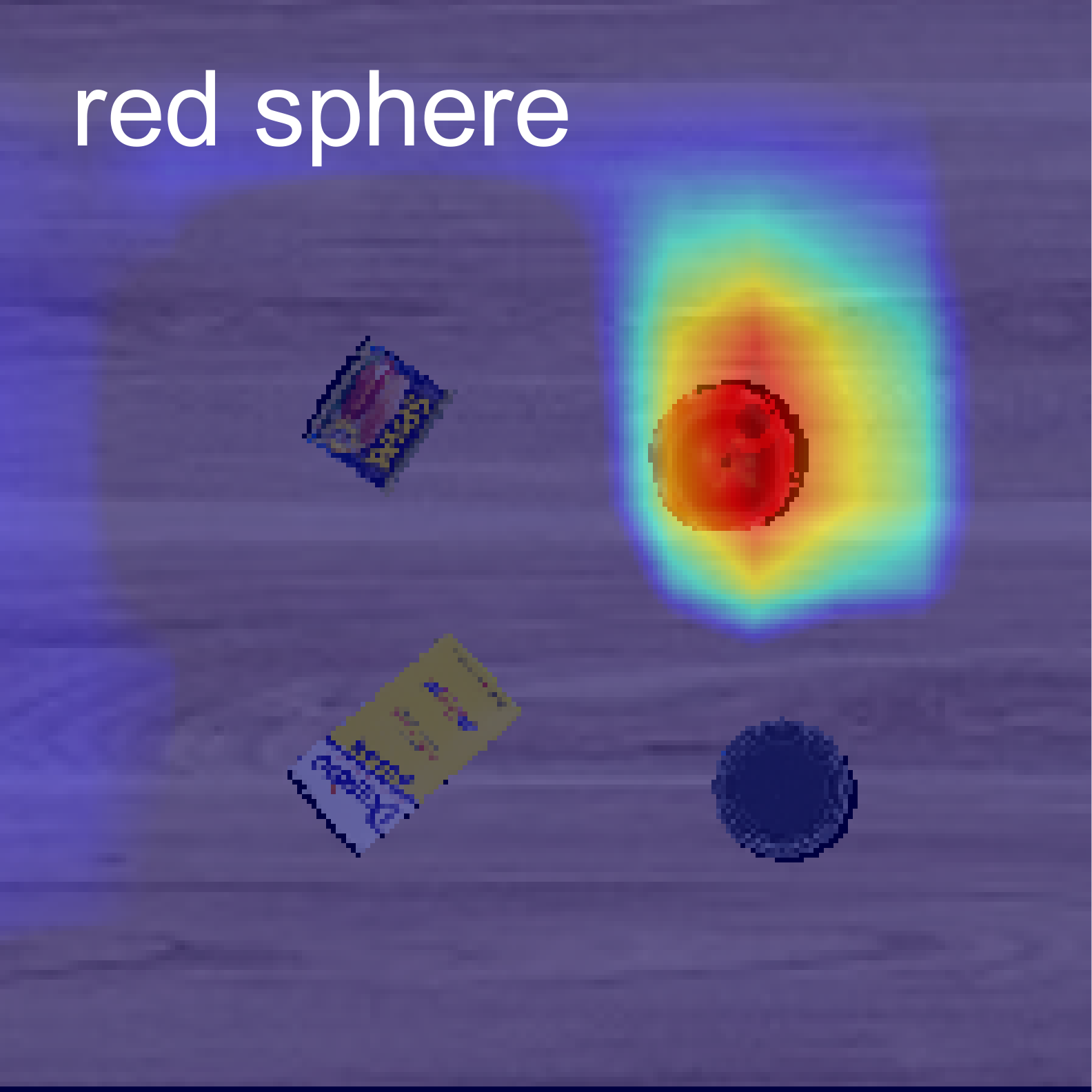}}\hfill
        {\includegraphics[width=0.24\textwidth]{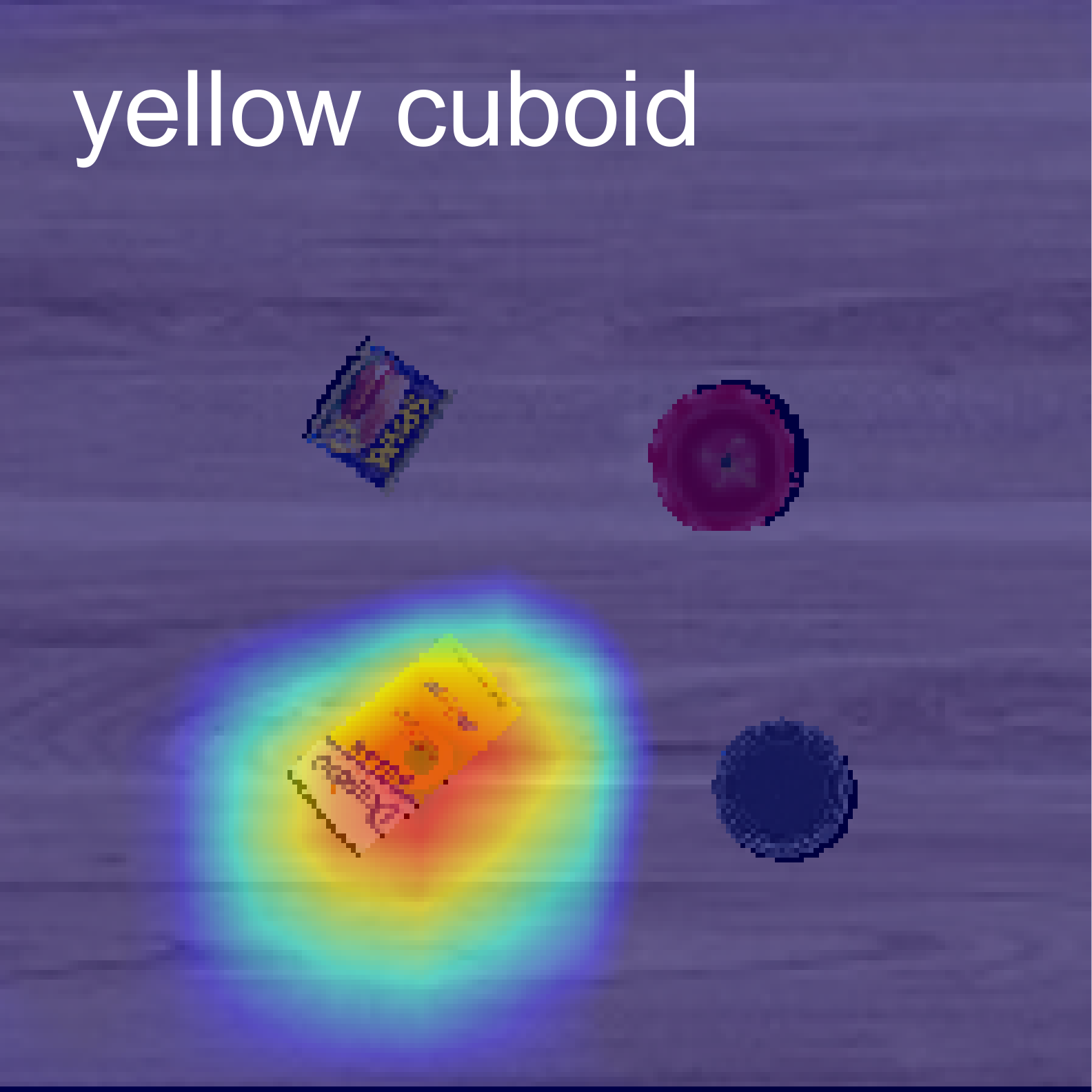}}\hfill
        {\includegraphics[width=0.24\textwidth]{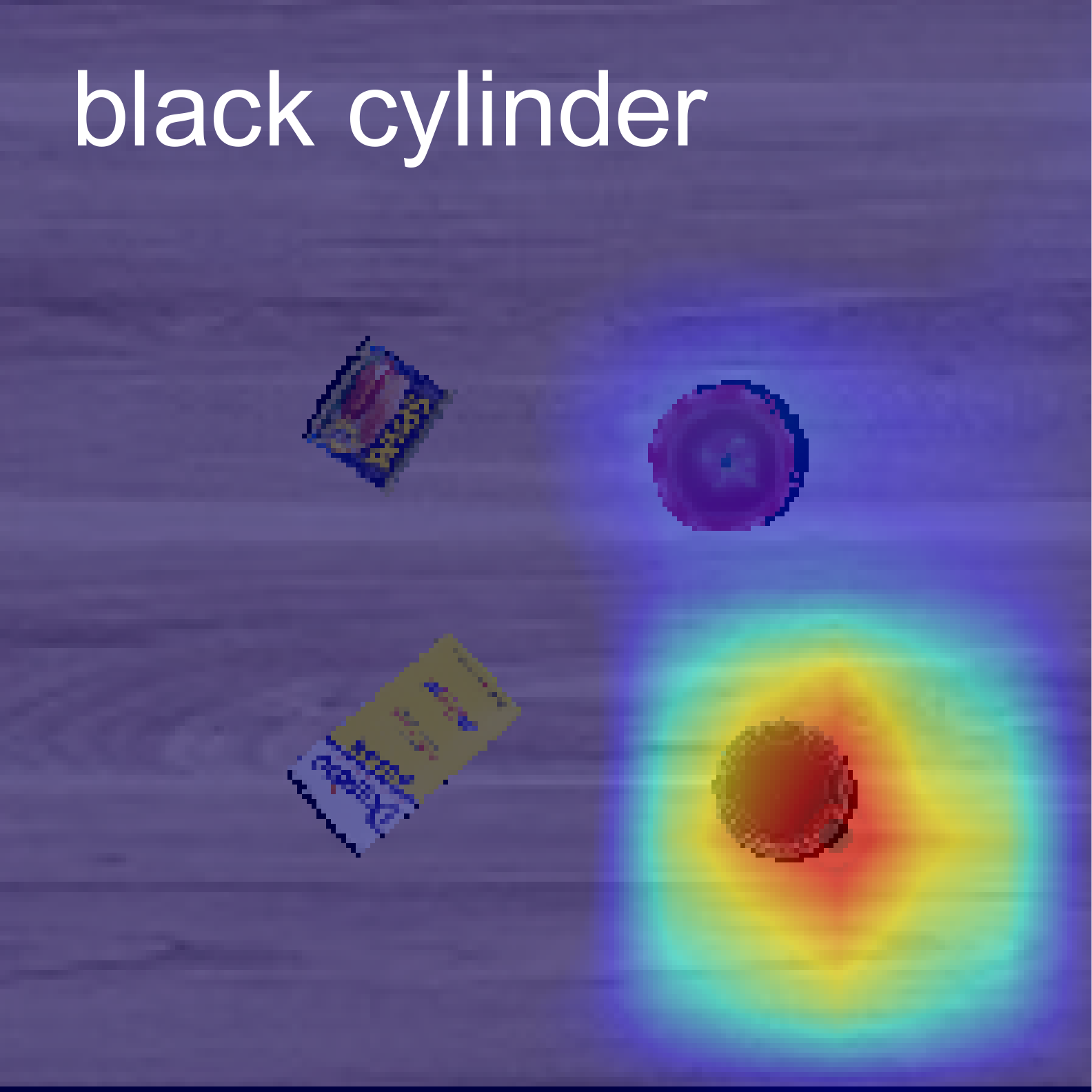}}\hfill
        {\includegraphics[width=0.24\textwidth]{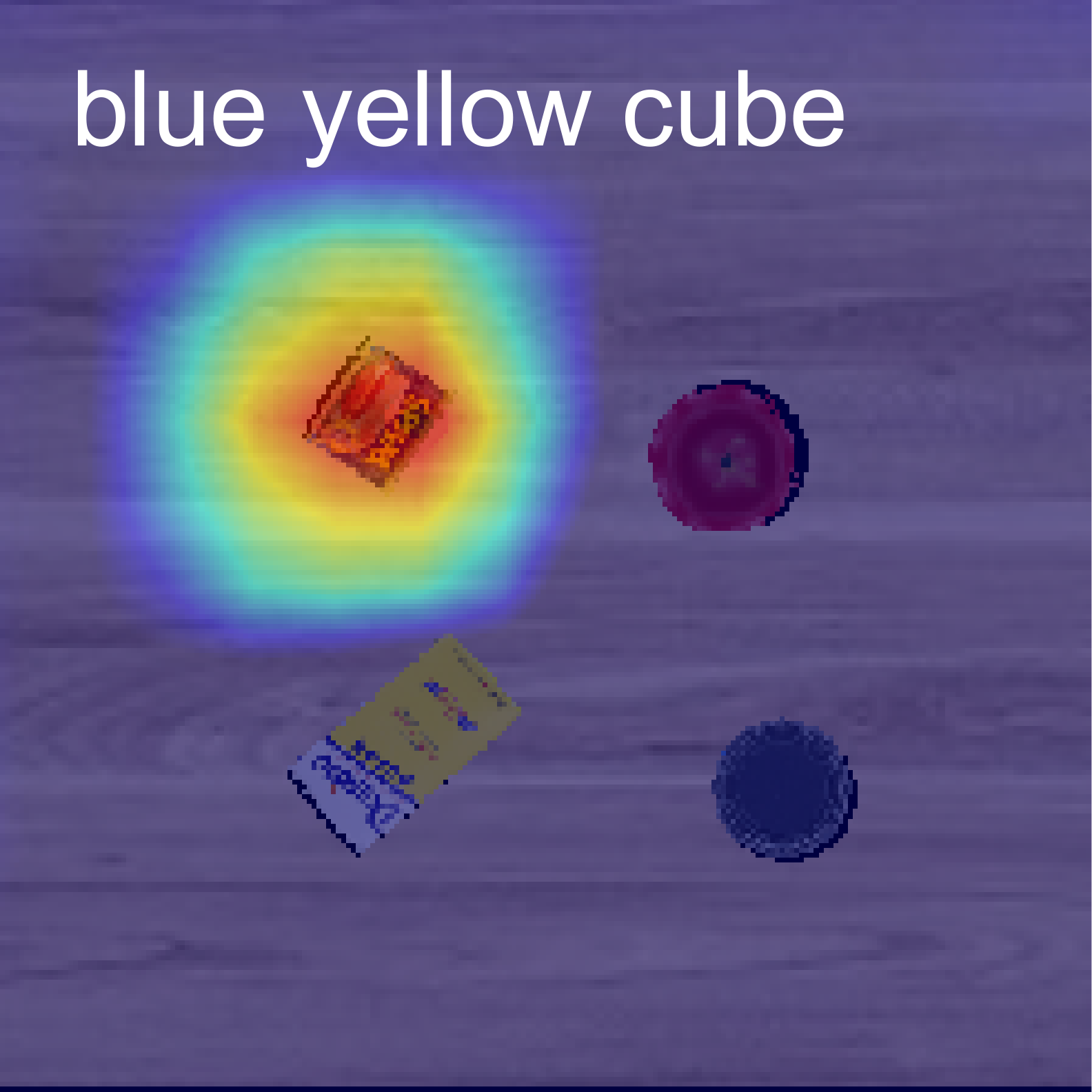}}%
        \vspace{-4pt}
        \caption{Encoder's attention}
        \vspace{2pt}
        \label{fig:metric}
    \end{subfigure}
    \begin{subfigure}{0.48\textwidth}
        {\includegraphics[width=0.24\textwidth]{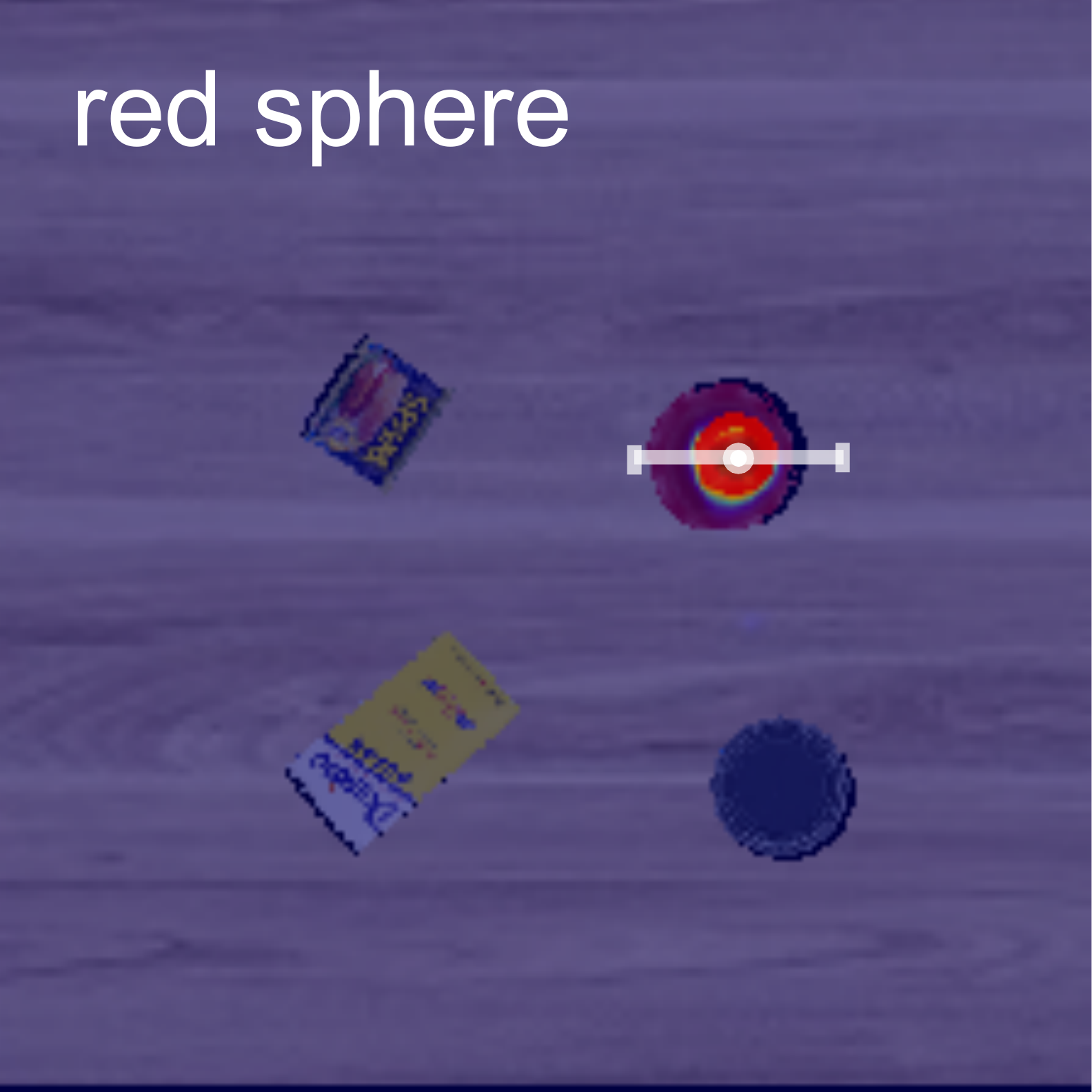}}\hfill
        {\includegraphics[width=0.24\textwidth]{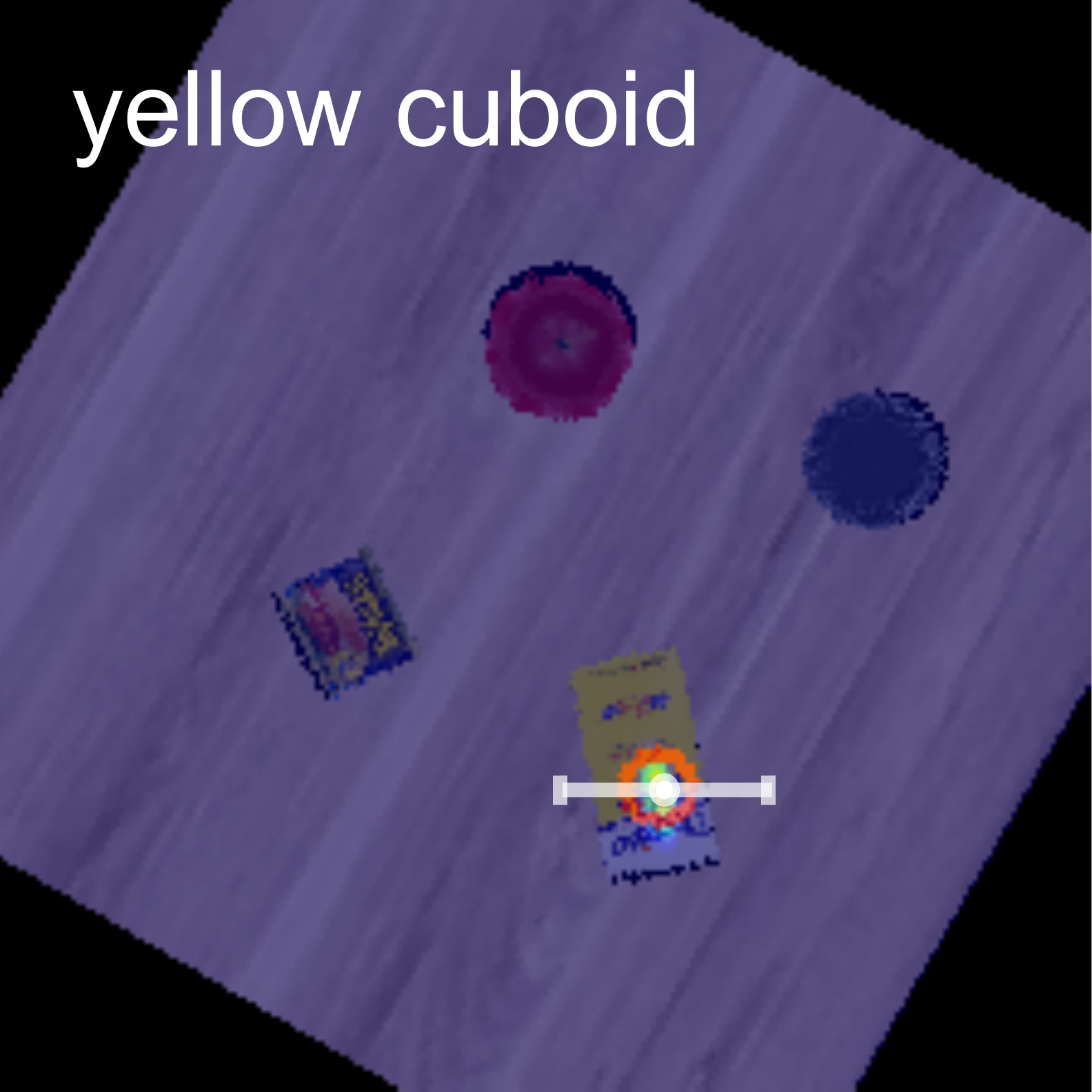}}\hfill
        {\includegraphics[width=0.24\textwidth]{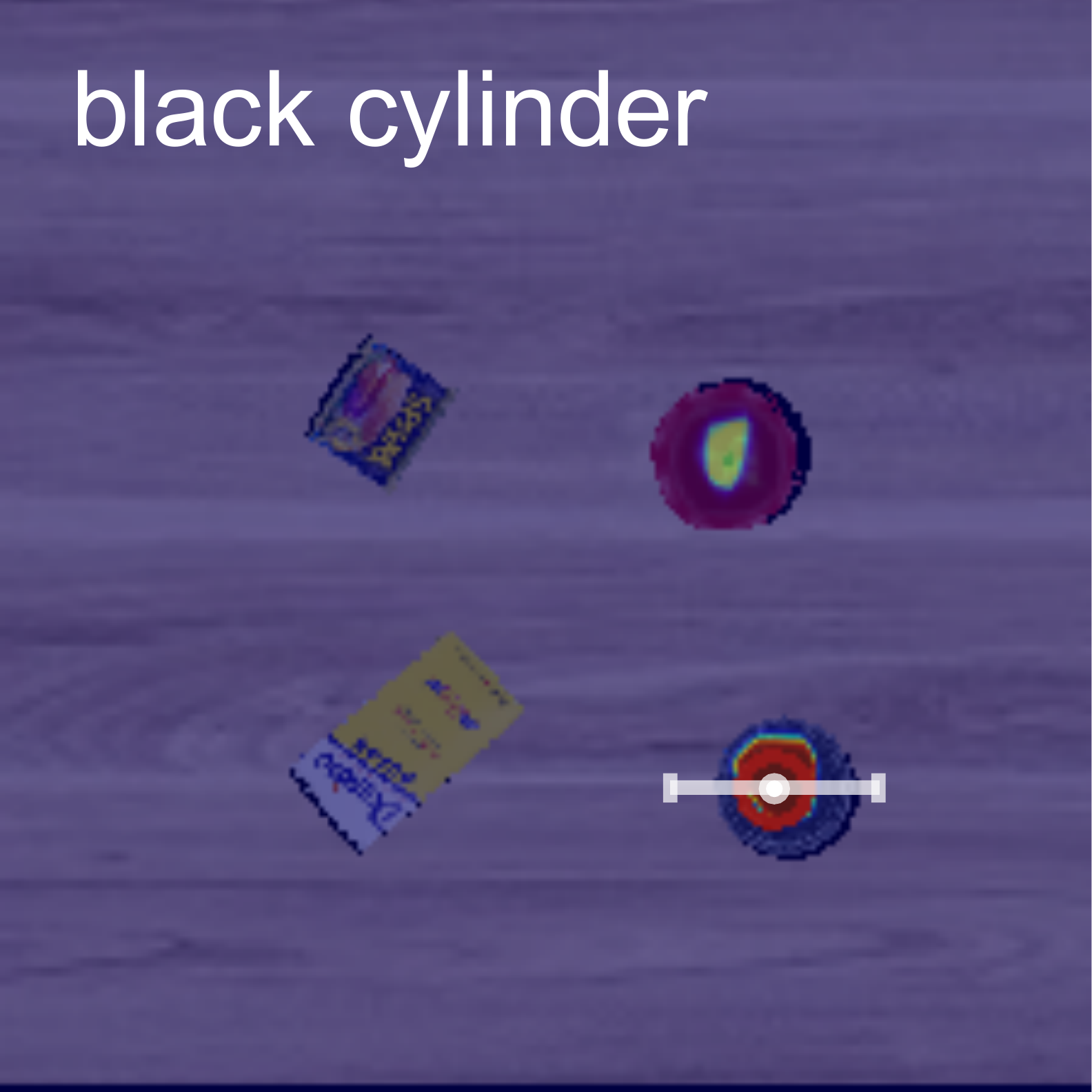}}\hfill
        {\includegraphics[width=0.24\textwidth]{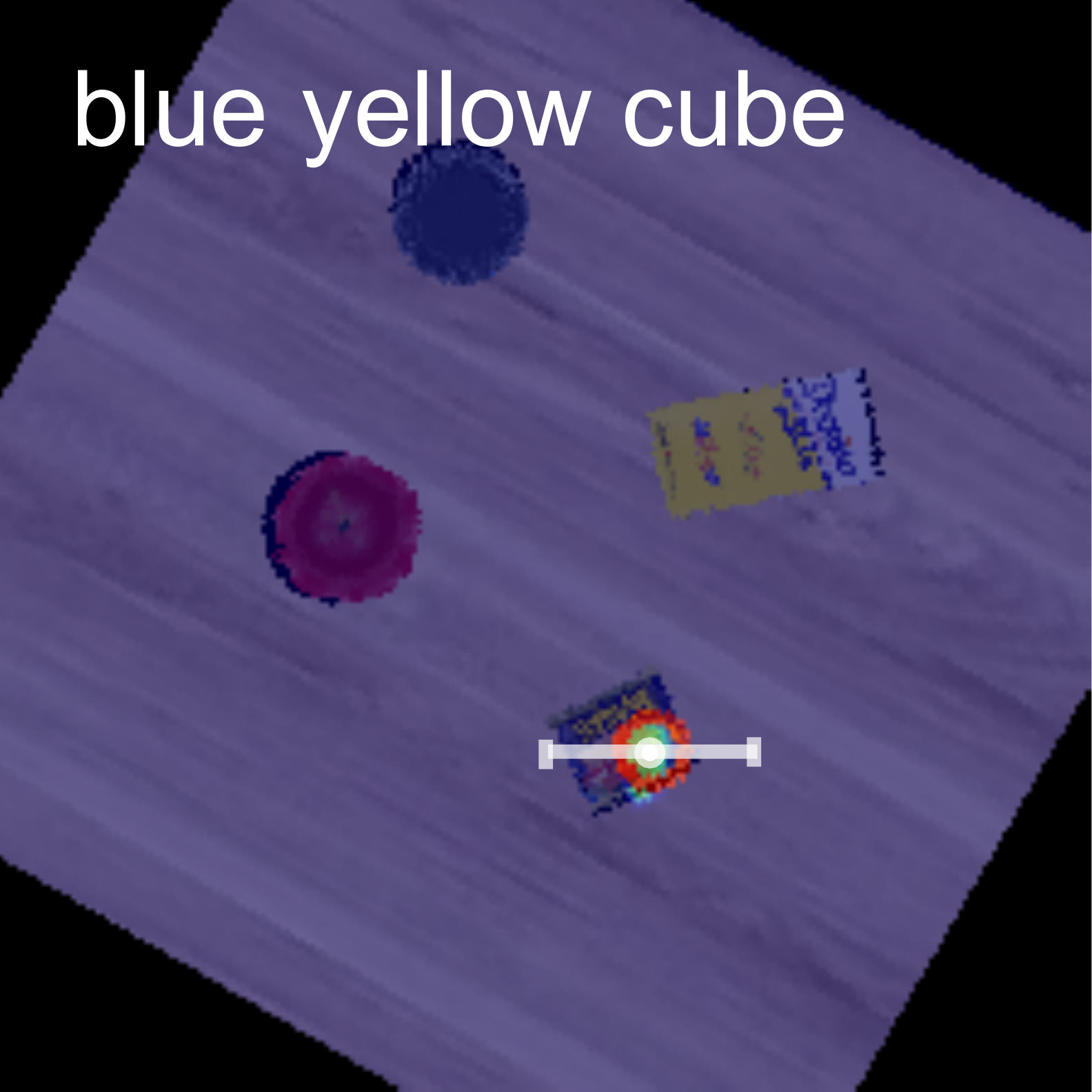}}%
        \vspace{-4pt}
        \caption{Grasping affordances}
        \vspace{-3pt}
        \label{fig:grasp}
    \end{subfigure}
    \caption{\textbf{Visualization of attention and grasping.} (a) shows attention of our encoder, and (b) shows heatmaps of grasping affordances for different targets (described by the query text). The maps of attention and grasping are consistent even for the novel objects.}%
    \label{fig:sim}
\end{figure}
\begin{table}[!b]
    \centering
    \caption{Target Recognition Accuracy (\%)}
    \label{tab:acc}
    \begin{tabular}{c|c|c|c}
    \hline
    Method & sim basic & sim novel & real\\
    \hline
    ClassIndis & 100.0 & 60.2 & 38.6\\
    EncoderIndis & 95.8 & 76.3 & 61.4 \\
    NoMetric & 93.4 & 75.0 & 58.3\\
    Ours-Attention & 95.3 & 79.4 & 75.8\\
    Ours & \textbf{100.0} & \textbf{84.5} & \textbf{77.3}\\
    \hline
    \end{tabular}
    \vspace{10pt} 
    \centering
    \caption{Instance Grasping Success Rate (\%)}
    \label{tab:grasp}
    \begin{tabular}{c|c|c|c}
    \hline
    Method & sim basic & sim novel & real\\
    \hline
    Indiscriminate & 27.2 & 25.8 & 15.2\\
    ClassIndis & 91.6 & 57.1 & 36.4\\
    EncoderIndis & 89.1 & 71.2 & 57.6 \\
    NoMetric & 90.5 & 70.7 & 55.3\\
    Ours & \textbf{98.4} & \textbf{80.3} & \textbf{69.7}\\
    \hline
    \end{tabular}
\end{table}

We evaluate the methods on both basic and novel objects in simulation, where there are 1200 test cases for the basic objects (Fig. \ref{fig:basic}) and 2000 test cases for the 20 novel objects (Fig. \ref{fig:testing_sim}, mostly from the YCB dataset \cite{calli2015ycb}). We assume the objects are placed right-side up to be stable while their 4D pose (3D position and a yaw angle) can vary arbitrarily. For each testing object, we pre-choose a query text that best describes its color and/or shape. In each test case, 4 objects are randomly sampled and placed in the workspace except avoiding any two objects with same attributes. The robot is required to grasp the target queried by an attribute text. We report the results of target recognition in Table \ref{tab:acc} and the results of instance grasping in Table \ref{tab:grasp}.
\begin{figure}[t]
    \centering
    \begin{subfigure}{0.48\textwidth}
        {\includegraphics[width=0.24\textwidth]{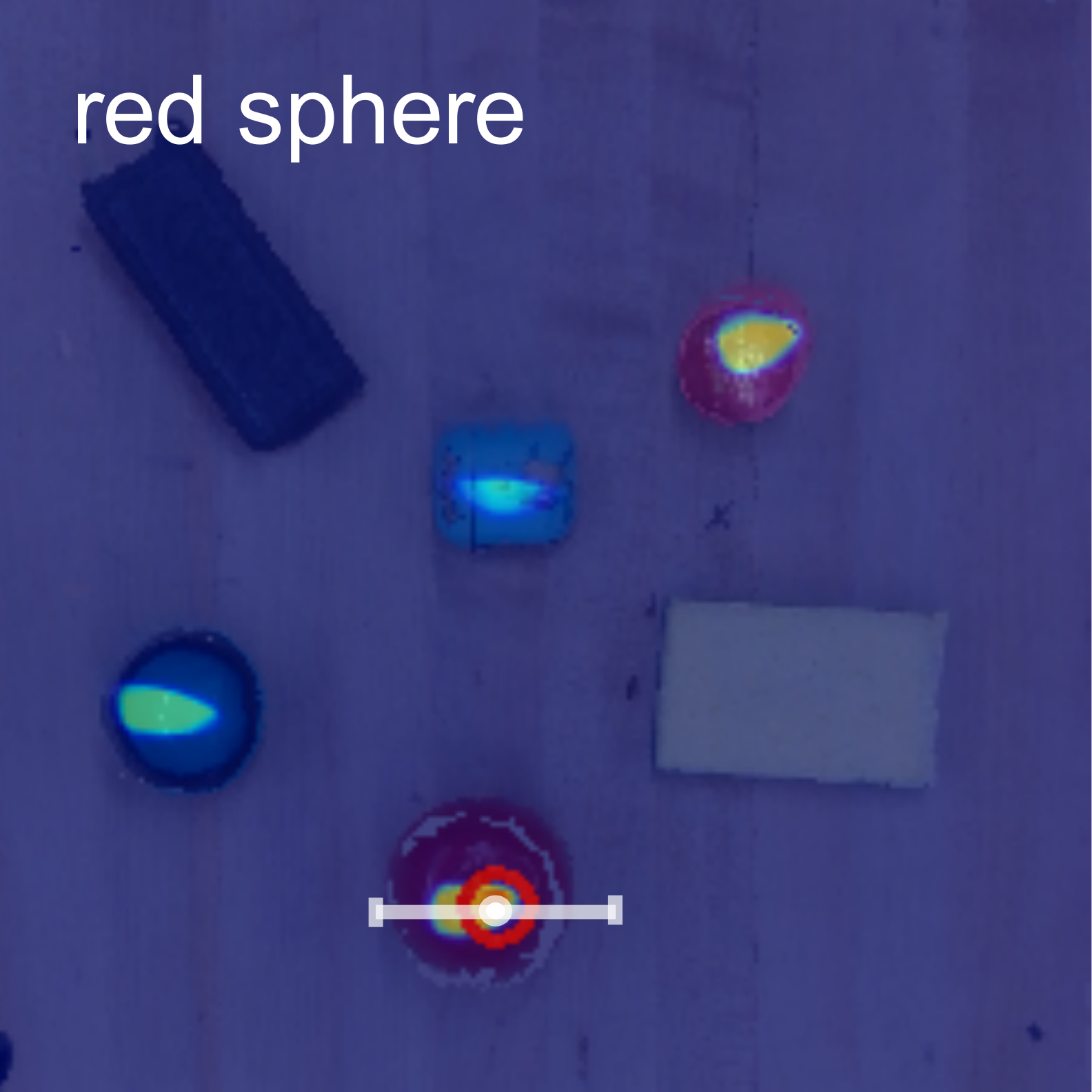}}\hfill
        {\includegraphics[width=0.24\textwidth]{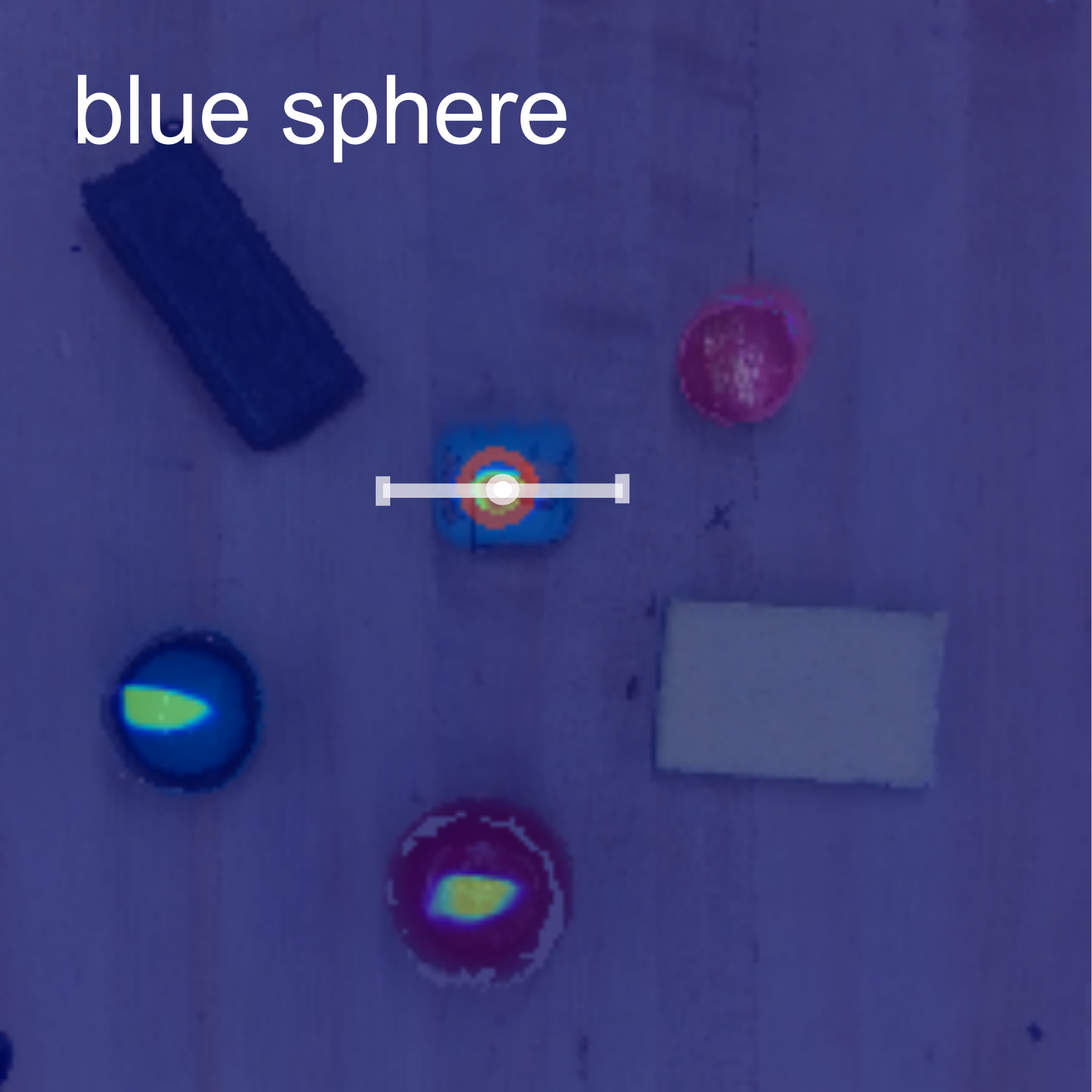}}\hfill
        {\includegraphics[width=0.24\textwidth]{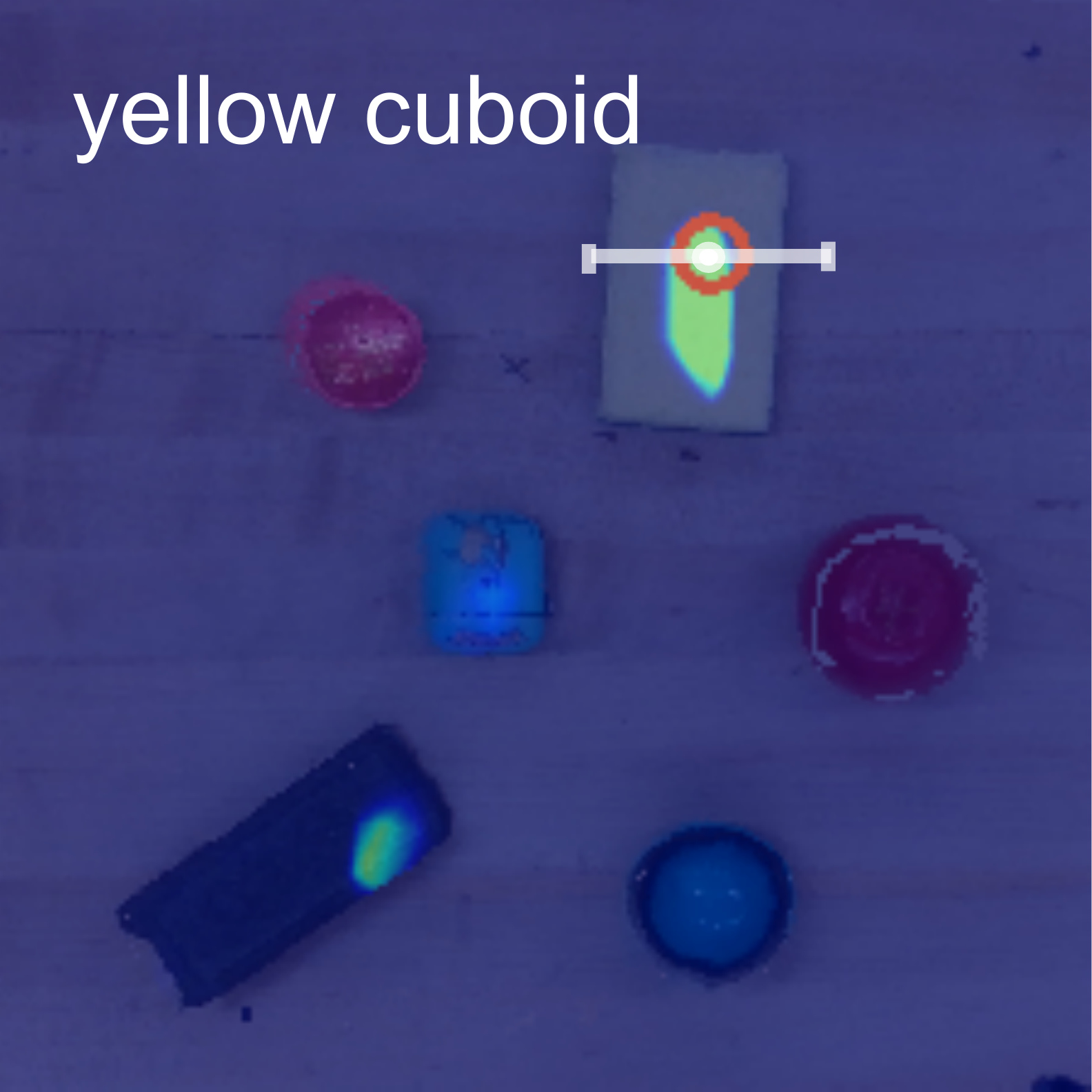}}\hfill
        {\includegraphics[width=0.24\textwidth]{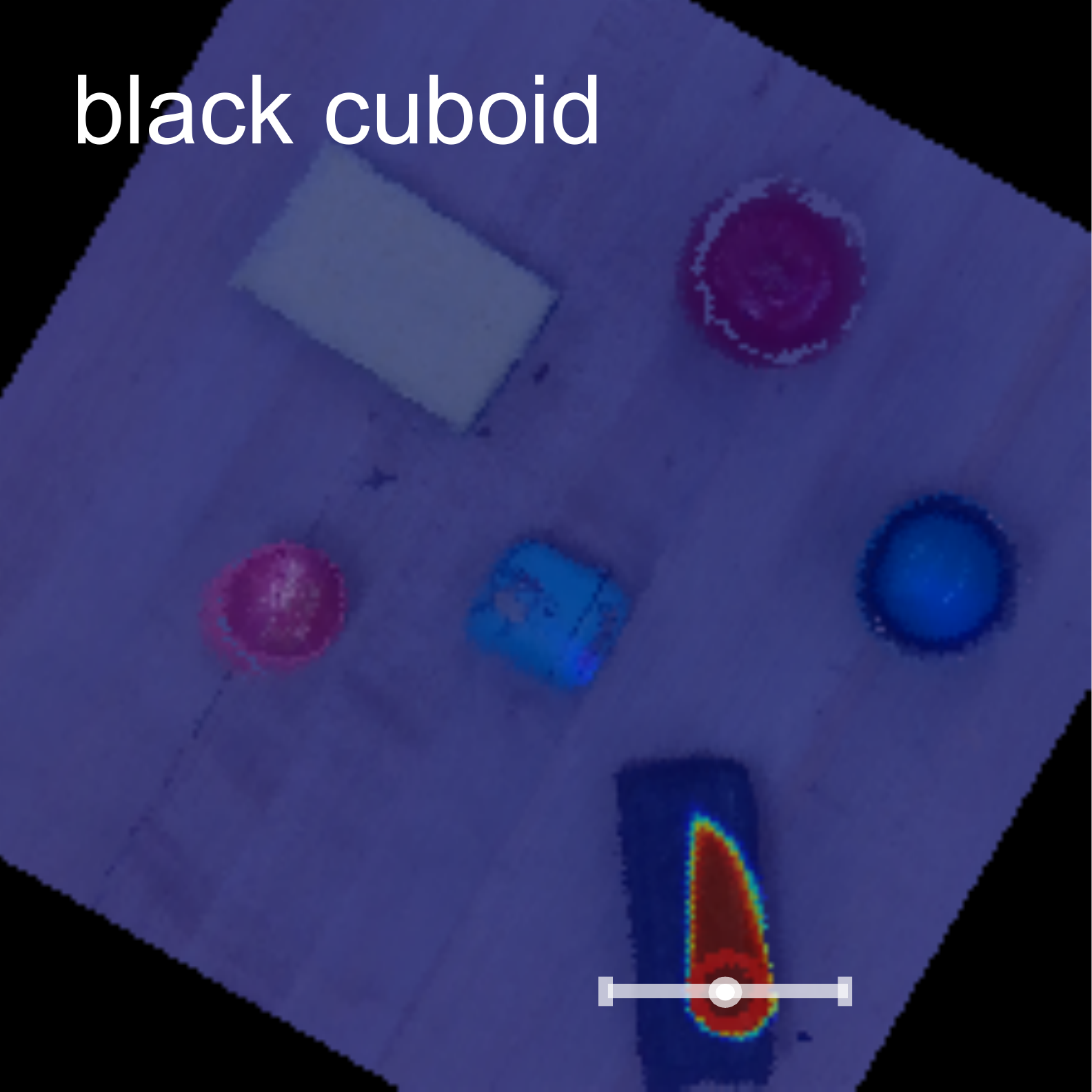}}%
        \vspace{-4pt}
        \caption{Generic model before adaptation}
        \vspace{2pt}
    \end{subfigure}
    \begin{subfigure}{0.48\textwidth}
        {\includegraphics[width=0.24\textwidth]{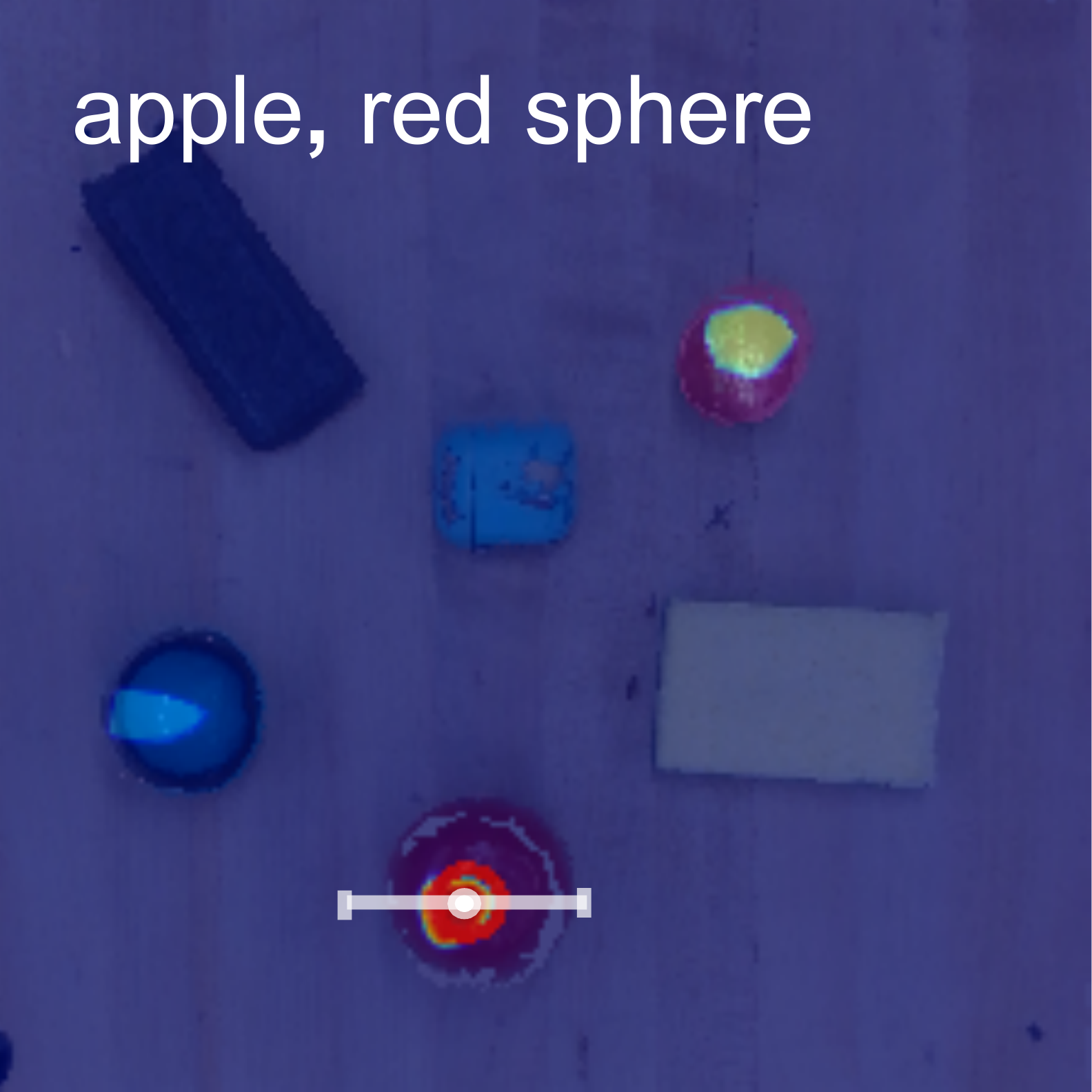}}\hfill
        {\includegraphics[width=0.24\textwidth]{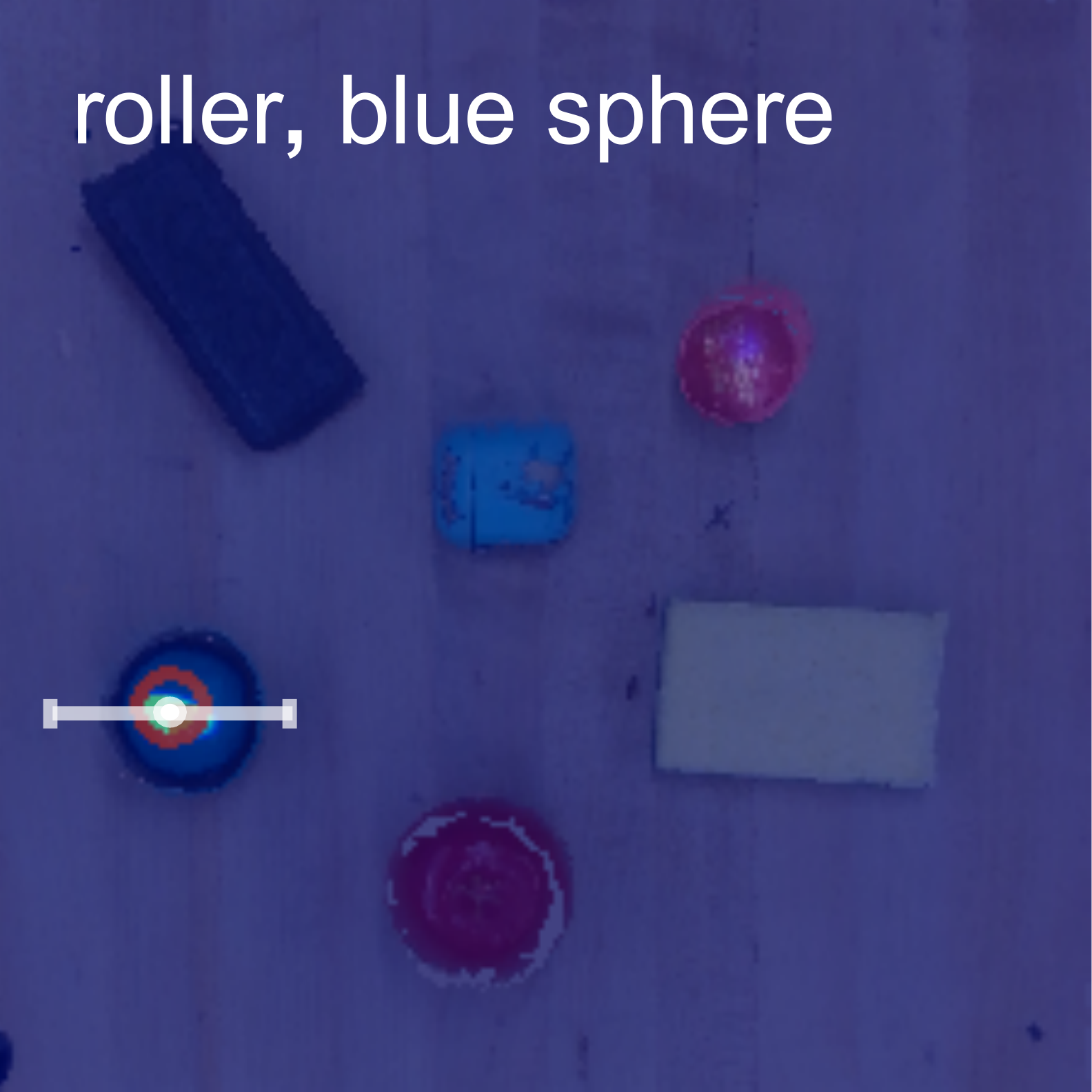}}\hfill
        {\includegraphics[width=0.24\textwidth]{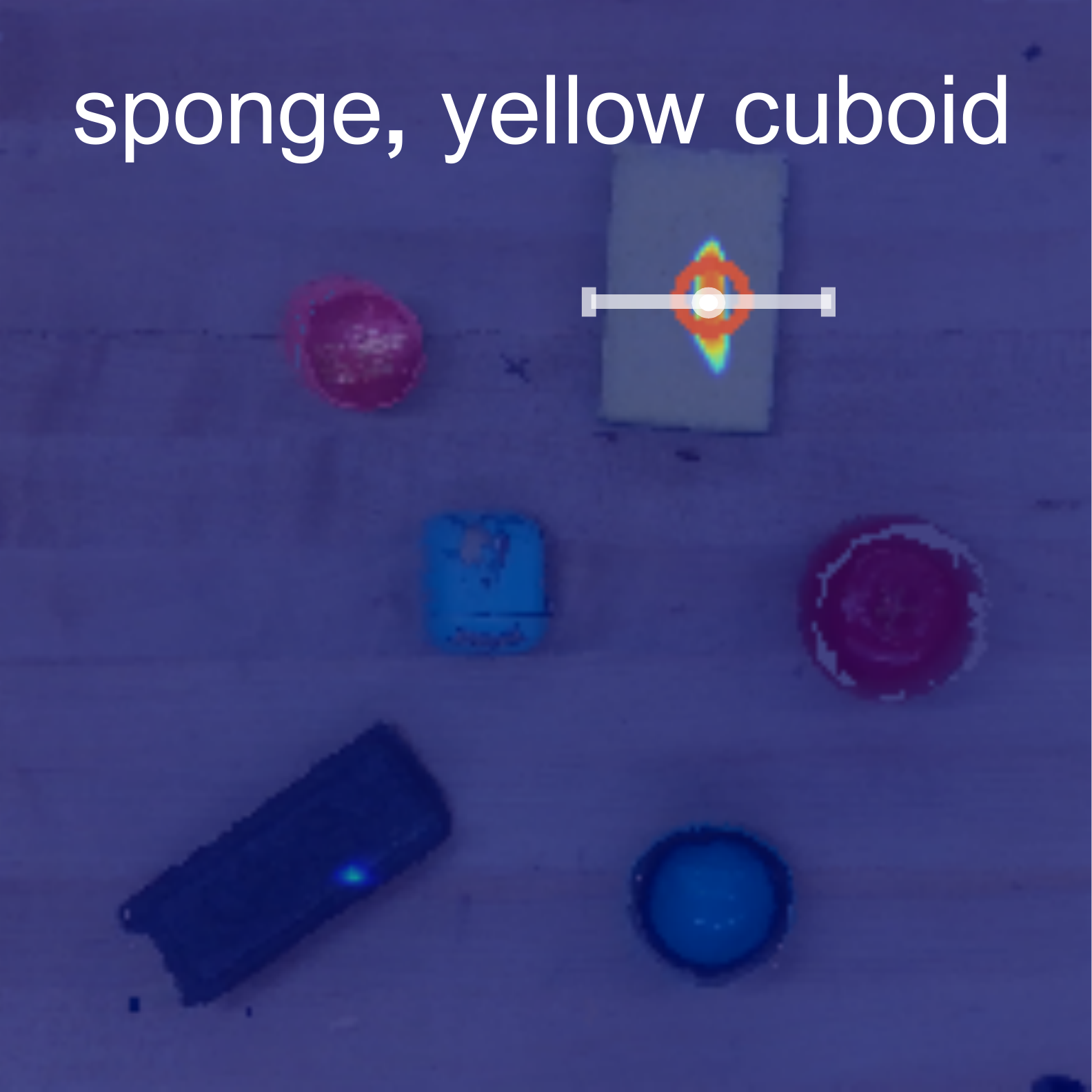}}\hfill
        {\includegraphics[width=0.24\textwidth]{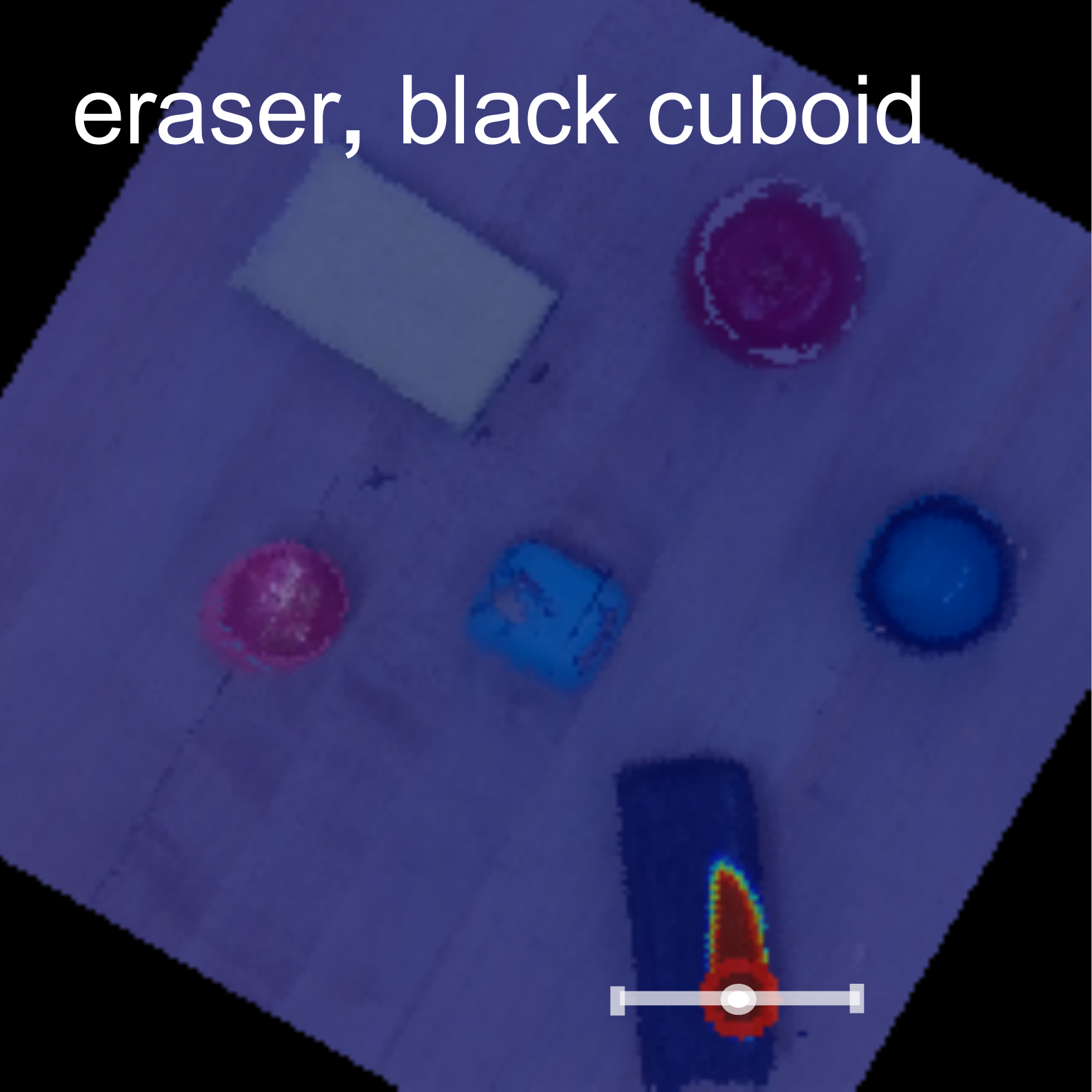}}%
        \vspace{-4pt}
        \caption{Adapted model after adaptation}
        \vspace{-3pt}
    \end{subfigure}
    \caption{\textbf{Visualization of grasping maps before and after adaptation.} (a) shows affordances from the generic model trained only with simulated basic objects, and (b) shows affordances from the adapted model after one-grasp adaptation. The adapted model improves in both recognition (e.g., roller) and grasping (e.g., sponge).}
    \label{fig:real}
    \vspace{-10pt}
\end{figure}
Overall, our approach outperforms the baselines remarkably (in both recognition and grasping) and achieves a 98.4\% grasping success rate on the simulated basic objects and an 80.3\% grasping success rate on the simulated novel objects. \textit{ClassIndis} extends \textit{Indiscriminate} that is well trained in target-agnostic tasks and performs well on the basic objects. But the attributes classifier generalizes poorly. \textit{EncoderIndis} utilizes a more generalizable recognition module and performs better on the novel objects. However, \textit{EncoderIndis} fails to reach optimality because its separately-trained recognition and grasping modules have different training objectives from instance grasping. As an ablation study, the performance gap between \textit{NoMetric} and \textit{Ours} shows the effectiveness of multimodal metric loss, which supervises the joint latent space to produce consistent embeddings, as discussed in Sec. \ref{subsec:exp_metric}. Our approach successfully learns object attributes that generalize well to novel objects, as shown in Fig. \ref{fig:sim}.

We further evaluate our approach and the baselines on the real robot before any adaptation (see Table \ref{tab:acc} and \ref{tab:grasp}). Fig. \ref{fig:testing_real} shows 11 testing objects of various colors and shapes used in our real-robot experiments. The robot is commanded to grasp the target within a combination of 6 objects placed on the table. We use the same 11 object combinations that are randomly generated and repeat each combination twice, resulting in a total of 132 grasping trials for each method. Overall, the grasping performance of all the methods decreases due to the sim2real gap. However, our approach shows the best generalization and achieves a 69.7\% grasping success rate, before adaptation, in the real-world scenes.
\subsection{One-Grasp Adaptation}
\label{subsec:exp_adapt}
Our generic model infers the closest object to the query text as the target and demonstrates good generalization despite the gaps in the testing scenes. Specifically, these gaps are 1) RGB values of the testing objects deviate from training ranges, 2) some testing objects are multi-colored, 3) shape and size differences between the testing objects and the training objects, and 4) depth noises in the real world cause imperfect object shapes. To account for the gaps, we further adapt our generic model to increase instance recognition and grasping performance of our approach.
\begin{table}[t]
    \centering
    \caption{Adapted Instance Grasping and Adaptation Gains (\%)}
    \label{tab:adapt}
    \begin{tabular}{c|c|c}
    \hline
    Method & sim novel & real\\
    \hline
    ClassIndis-Adapt & 59.8 / 2.7 & 39.4 / 3.0\\
    EncoderIndis-Adapt & 75.3 / 4.1 & 62.9 / 5.3\\
    NoMetric-Adapt & 71.8 / 1.1 & 56.1 / 0.8\\
    Ours-Adapt & \textbf{87.6} / \textbf{7.3} & \textbf{80.3} / \textbf{10.6}\\
    \hline
    \end{tabular}
    \vspace{-10pt}
\end{table}

We first collect one successful grasp of a solely placed target object and then augment the collected data by rotating with additional $N-1$ angles, as discussed in Sec. \ref{subsec:method_adapt} and shown in Fig. \ref{fig:adapt}. As a comparison, the baseline methods are also adapted with the same adaptation data: 1) \textbf{\textit{ClassIndis-Adapt}} updates its attributes classifier for a better recognition accuracy on the adaptation data. 2) \textbf{\textit{EncoderIndis-Adapt}} minimizes the latent distance between cropped target images and query text to improve text template matching. 3) \textbf{\textit{NoMetric-Adapt}} takes as input images and text, and minimizes motion loss on the adaptation data. We also update \textit{Indiscriminative} grasping in \textit{ClassIndis-Adapt} and \textit{EncoderIndis-Adapt}.

We keep the experimental setup the same with Sec. \ref{subsec:exp_grasp} and evaluate the instance grasping performance of the adapted models, as reported in Table \ref{tab:adapt}. The attributes classifier in \textit{ClassIndis-Adapt} suffers from insufficient adaptation data, limiting its recognition and adaptation performance. While \textit{EncoderIndis-Adapt} minimizes embedding distances in its latent space and shows superior performance, it is still worse than \textit{Ours-Adapt}. By fine-tuning over the structured metric space, \textit{Ours-Adapt} updates the end-to-end model and improves recognition and grasping jointly (see Fig. \ref{fig:real}). At the cost of minimal data collection and quick adaptation, \textit{Ours-Adapt} achieves an 87.6\% grasping success rate on the simulated novel objects and an 80.3\% grasping success rate on the real objects, which shows the significant adaptation gains. On the contrary, the unstructured latent space in \textit{NoMetric-Adapt} limits its adaptation, demonstrating the importance of attributes learning for grasping affordances learning.
\section{CONCLUSION}
In this work, we presented a novel attribute-based robotic grasping system. An end-to-end architecture was proposed to learn object attributes and manipulation jointly. Workspace images and query text are encoded into a joint metric space, which is further supervised by object persistence before and after grasping. Our model was self-supervised in a simulation only using basic objects but showed generalization with one-grasp adaptation capability to novel objects and real-world scenes. Our grasping system achieved an $87.6\%$ instance grasping success rate in simulation and an $80.3\%$ instance grasping success rate in the real world, both on unknown objects. Our approach outperformed the other compared methods by large margins. 


\bibliographystyle{IEEEtran}
\bibliography{references}

\clearpage
\section{APPENDIX}
\includepdf[offset=0 -0.6cm]{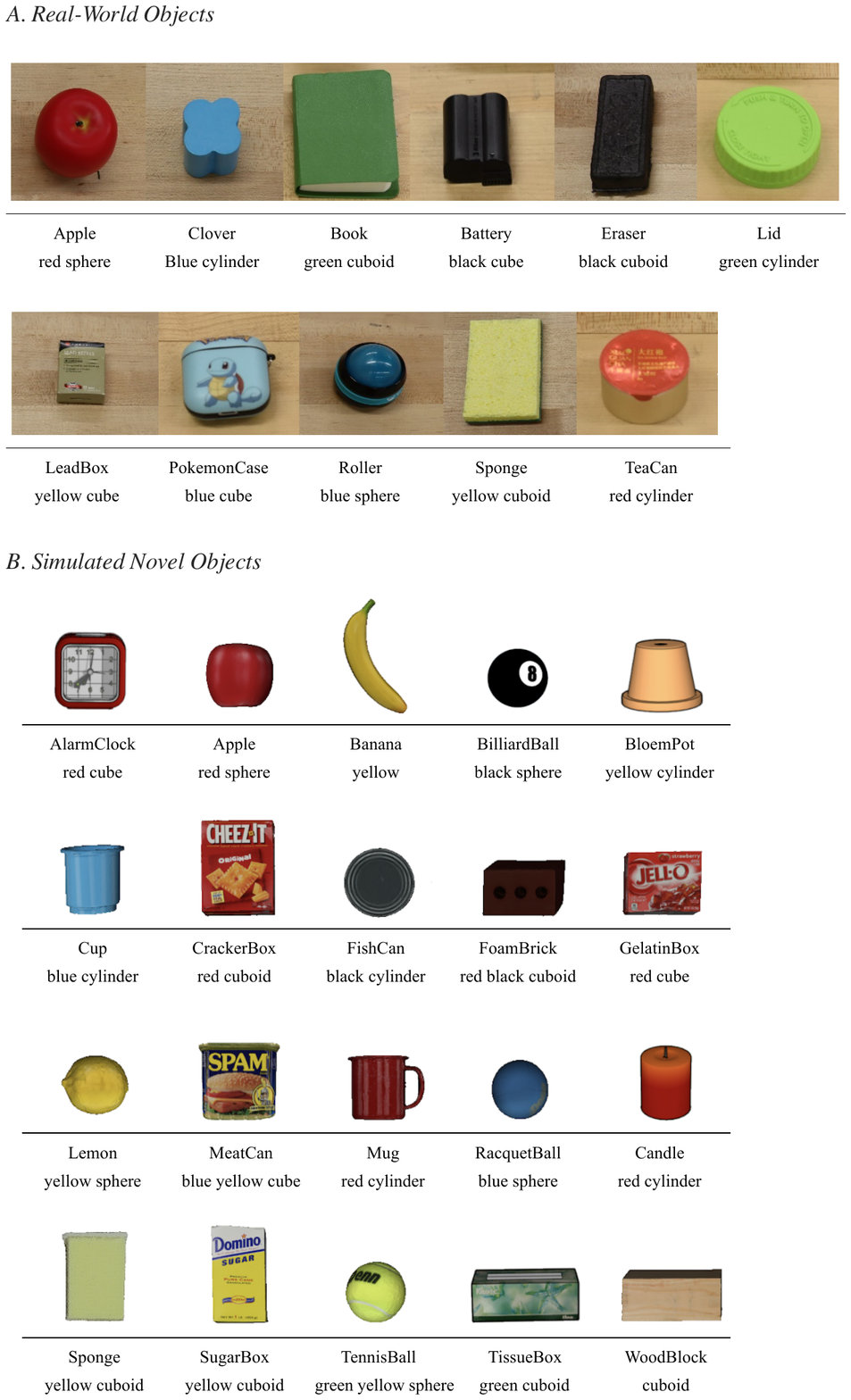}

\end{document}